\documentclass{tlp}

\usepackage{amsmath}
\usepackage{amssymb}
\usepackage{graphicx}
\usepackage{multirow}

\usepackage{xspace}
\usepackage{hyperref}
\usepackage{url}
\usepackage{tikz}
\usepackage{caption}
\usepackage{subcaption}
\usepackage{listings}
\usepackage{verbatimbox}

\def\mapf{{MAPF}\xspace}
\def\dmapf{{D-MAPF}\xspace}

\def\ii#1{\hbox{\it #1\/}}

\def\seq#1{\left\langle #1 \right\rangle}

\def\eqs{\,{=}\,}
\def\neqs{\,{\neq}\,}
\def\leqs{\,{\leq}\,}
\def\geqs{\,{\geq}\,}
\def\lts{\,{<}\,}
\def\gts{\,{>}\,}
\def\ins{\,{\in}\,}
\def\cups{\,{\cup}\,}
\def\caps{\,{\cap}\,}
\def\setminuss{\,{\setminus}\,}
\def\pluss{\,{+}\,}
\def\subseteqs{\,{\subseteq}\,}
\def\downarrs{{\downarrow}}
\def\uparrs{{\uparrow}}

\newcommand{\event}[1]{\seq{A_{#1}\uparrs, A_{#1}\downarrs, O_{#1}\uparrs, O_{#1}\downarrs, {#1}}}

\def\trav#1{{r_{a_{#1}}^{P_{#1}} \eqs \seq{x_{#1}, y_{#1}, f_{#1}}}} 
\def\travp#1{{r_{a_{#1}}^{P'_{#1}} \eqs \seq{x_{#1}, y_{#1}, f_{#1}}}} 
\def\travi#1{r_{a_{#1}}^{P_{#1}}} 
\def\travcoll{\mathbf{r}_A^\mathbf{P}} 

\usepackage{todonotes}

\begin{document}

\lefttitle{Bogatarkan and Erdem}

\jnlPage{}{}
\jnlDoiYr{2025}
\doival{}

\title[A General Framework for Dynamic MAPF]{A General Framework for Dynamic MAPF \\ using Multi-Shot ASP and Tunnels}

\begin{authgrp}
\author{\sn{Aysu} \gn{Bogatarkan} \and \sn{Esra} \gn{Erdem}}
\affiliation{Sabanci University, Faculty of Engineering and Natural Sciences, Istanbul, Turkiye \\ \emails{aysubogatarkan@sabanciuniv.edu}{esraerdem@sabanciuniv.edu} }
\end{authgrp}


\maketitle

\begin{abstract}
    \mapf problem aims to find plans for multiple agents in an environment within a given time, such that the agents do not collide with each other or obstacles. Motivated by the execution and monitoring of these plans, we study Dynamic \mapf (\dmapf) problem, which allows changes such as agents entering/leaving the environment or obstacles being removed/moved. Considering the requirements of real-world applications in warehouses with the presence of humans, we introduce 1) a general definition for \dmapf (applicable to variations of \dmapf), 2) a new framework to solve \dmapf (utilizing multi-shot computation, and allowing different methods to solve \dmapf), and 3) a new ASP-based method to solve \dmapf (combining advantages of replanning and repairing methods, with a novel concept of tunnels to specify where agents can move). We have illustrated the strengths and weaknesses of this method by experimental evaluations, from the perspectives of computational performance and quality of solutions.
\end{abstract}



\section{Introduction}
\label{sec:intro}

The multi-agent path finding (\mapf) problem aims to find plans for multiple agents in a shared environment such that the agents do not collide with each other or obstacles, subject to constraints on the total/maximum plan length. 
%
The dynamic \mapf (\dmapf) problem considers the changes that occur in the environment or in the team during the execution of a \mapf plan (e.g., new agents joining the team, existing agents leaving the environment, new obstacles being added, existing obstacles being removed or moved) and aims to find a new plan for agents to reach their goals.




The existing approaches 
to solving \dmapf consider different objective functions (e.g., minimizing the makespan or the sum of costs) and changes (e.g., team or environment changes), make different assumptions on entrances and departures of agents (e.g., agents immediately or later appear at their initial locations, agents stay or disappear when they reach their destinations), and investigate different methods to solve \dmapf. 

For instance, Lifelong \mapf~\citep{wan2018lifelongdynamic,Li2021lifelongwarehouse} considers assignment of new goal locations to the agents who have completed their plans, which can be viewed as adding a new agent. 
An incremental version of CBS algorithm was introduced by \cite{wan2018lifelongdynamic} such that solutions for new agents are computed while re-using the solutions for the existing agents and they are adjusted only if needed at each time step. 
The idea of Windowed \mapf was used by \cite{Li2021lifelongwarehouse}, where the instance is splitted into a sequence of instances, replanning is done periodically after a specific number of steps, and the collisions are resolved only for the steps within a given window. 

Online \mapf~\citep{svancara2019} considers the addition of new agents as a dynamic change, and allows the agents to appear from a garage after some waiting and disappear from the environment once it reaches its goal. Different methods were introduced for solving this problem with SAT based and search based solvers and these methods are further investigated by \cite{morag2022online} in terms of their quality. One of the methods introduced for solving Online \mapf is the \textit{Replan-All} method, that discards the existing plan and re-solves \mapf for all agents considering the changes. \cite{Ma2021online} investigates Online \mapf problem theoretically, gives complexity results and provides a classification for online \mapf algorithms. \cite{Ho2019uav} consider an online and 3-dimensional setting, with heterogeneous agents being added during execution and solves the problem using modified versions of CBS and ECBS algorithms. 

Different from these studies, in terms of changes, \cite{bogatarkanPE19} consider leaving agents and changes in obstacles as well. Their \textit{Revise-and-Augment} method reuses the existing plan: when a change occurs, the plans of existing agents are revised by rescheduling their waiting times, while plans are computed for the new agents. \cite{Atiq_2020} consider a similar problem. 
When a change occurs in the environment, a minimal subset of agents having conflicts are identified and replanning is applied to resolve conflicts.

In this study, we investigate \dmapf further with the following motivations, and theoretical and practical contributions.

$\bullet$~
To be able to study rich variations of \dmapf mentioned above, on a common ground, we introduce a rigorous definition for \dmapf, that is general enough to cover 1)~various changes in the environment and the team of agents over time, 2)~different objective functions on plans, and 3)~different assumptions on appearances/disappearances of agents, and that is not specifically oriented towards a particular method.

$\bullet$~
We introduce a new framework to solve \dmapf, that is general and flexible enough to allow different replanning and/or repairing methods. With the motivation of a modular architecture and efficient computations, our framework utilizes multi-shot computation~\citep{gebser2019multi} based on  ASP~\citep{lifschitz08,GelfondL91}.
Recall that multi-shot solving allows changes to the input ASP program in time, by introducing an external control to the ASP system. The external control allows operations, such as adding and grounding new programs, assigning truth values of some atoms, and solving the updated program, while the ASP system is running. Our earlier studies \citep{bogatarkanPE19} utilize single-shot computation.

$\bullet$~
We design and implement the Replan-All and Revise-and-Augment methods using multi-shot ASP, and integrate them in the general \dmapf framework. We empirically observe that multi-shot Replan-All is computationally more efficient but sometimes dramatic changes in the paths of the existing agents occur in the recomputed plans. Such changes are not desired from the perspective of real-world applications. For instance, in a warehouse where robots collaborate with human workers,
changes in the routes of robots might be unexpected, distracting, unsafe, and inefficient for human workers. 

$\bullet$~
We introduce a new method for \dmapf, called \textit{Revise-and-Augment-in-Tunnels}, that combines the advantages of these two methods. Unlike revise-augment, this method does not require that every existing agent follow their existing paths while revising their plans. Instead, 1) it  creates a ``tunnel'' for each existing agent, that consists of the agent's existing path and the neighboring locations within a specified ``width'', and 2) it allows every existing agent to follow a path within their own tunnel while it revises their plans. So the existing agents do not have to follow their previously computed paths. While revising the plans of existing agents within their tunnels, 3) the Revise-and-Augment-in-Tunnels method computes plans for the new agents and augments these plans with the revised plans, respecting the collision constraints. Note that as the tunnel width gets larger (resp. smaller), the Revise-and-Augment-in-Tunnels method gets closer to the Replan-All method (resp. the Revise-and-Augment method).

$\bullet$~
We implement the Revise-and-Augment-in-Tunnels method using multi-shot ASP, and integrate it in our \dmapf framework. We design and perform experiments to better understand the strengths and the weaknesses of this new method, considering computational performance (in time) and quality of solutions (in terms of plan changes).


\section{Preliminaries: Paths, Traversals, Plans}
\label{sec:prel}

We introduce a more general definition for \mapf problem, compared to the earlier definitions~\citep{ErdemKOS13,SternSFK0WLA0KB19}, so that it allows us to explicitly state our assumptions and consider different cost functions depending on the particular application. 

Let us first introduce the relevant concepts and notations. Consider an undirected graph $G \eqs (V,E)$. 
A path $P \eqs \seq{v_0,\dots,v_n}$ in $G$ is a sequence of vertices $v_i \ins V$ such that, for every $\seq{v_i,v_{i\pluss 1}}$ in $P$, there exists an edge $\{v_i,v_{i\pluss 1}\} \ins E$.


Let $A$ be a set of agents. Every agent $a_i \ins A$ is characterized by an initial location $init_i \ins V$, a goal location $goal_i \ins V$, and a joining time $join_i$ ($join_i\geqs 0$).


For every agent $a_i$ and every path $P_i$, a {\em traversal} $r_{a_i}^{P_i}$ of path $P_i$ by agent $a_i$ is characterized by a starting time $x_i$ ($x_i\geqs 0$), an ending time $y_i$ ($y_i\geqs x_i)$, and a function~$f_i$ that maps every integer $t$ ($x_i \leqs t \leqs y_i$) to a vertex in $P_i$, such that, for every $v_k, v_{k\pluss 1}$ in $P_i$ and, for every $t$, if $f_i(t) \eqs v_k$ then $ f_i(t\pluss 1) \eqs v_k$ or $f_i(t\pluss 1) \eqs v_{k\pluss 1}$.
We denote by $\mathbf{P}_A$ the collection of paths for every agent in $A$, and by $\mathbf{r}_A^\mathbf{P}$ the collection of traversals $\trav{i}$ of every agent $a_i \ins A$.

For two agents $a_i, a_j$ with traversals $\trav{i}$ and $\trav{j}$, we say that
$a_i$ and $a_j$
    collide at a vertex at time $t$, if for some $t$ where $\max(x_i, x_j) \leqs t \leqs \min(y_i, y_j)$, $f_i(t) \eqs f_j(t)$. 
  We say that  $a_i$ and $a_j$
  collide at an edge between times $t$ and $t\pluss 1$, if for some $t$ where $\max(x_i,x_j) \leqs t \lts \min(y_i,y_j)$, $f_i(t) \eqs f_j(t\pluss 1)$ and $f_i(t\pluss 1) \eqs f_j(t)$.
These types of collisions are called {\em vertex conflicts} and {\em swapping conflicts}, respectively.

\begin{figure}[t]
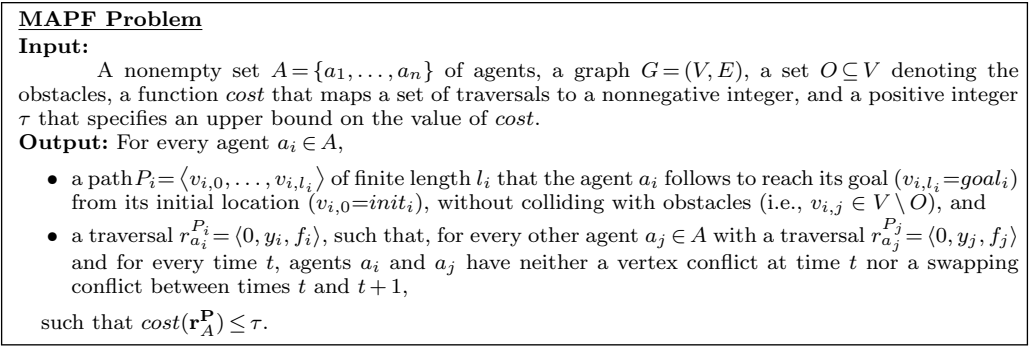

    \centering%
    {
    \fbox{
    \begin{minipage}{0.98\columnwidth}
        \footnotesize
        {\bf\underline{\mapf Problem}}

        \vspace{.25ex}
        {\bf Input:} 

        \hspace{9mm} A nonempty set $A \eqs \{a_1,\dots ,a_n\}$ of agents,
        a graph $G\eqs(V,E)$,
        a set $O\subseteqs V$ denoting the obstacles,
        a function $cost$ that maps a set of traversals to a nonnegative integer, and
        a positive integer $\tau$ that specifies an upper bound on the value of $cost$.

        {\bf Output:} For every agent $a_i \ins A$,
        \begin{itemize}[leftmargin=7mm]
            \item[$\bullet$]
             a path $\!P_i \!\eqs\! \seq{v_{i,0}, \dots, v_{i,l_i}}$ of finite length $l_i$ 
                that the agent $a_i$ follows to reach its goal ($v_{i,l_i} \!\eqs\! goal_{i}$) from its initial location ($v_{i,0} \!\eqs\! init_{i}$),
                without colliding with obstacles (i.e., $v_{i,j} \in V \setminuss O$), and

            \item[$\bullet$] a traversal $\travi{i}{=}\seq{0,y_i,f_i}$, such that, for every other agent $a_j\ins A$ with a traversal $\travi{j}{=}\seq{0,y_j,f_j}$ and for every time $t$, agents $a_i$ and $a_j$ have
                neither a vertex conflict at time $t$
                nor a swapping conflict between times $t$ and $t\pluss 1$,
        \end{itemize}%
        \hspace{2mm} such that $cost(\travcoll) \leqs \tau$. 
    \end{minipage}}}

    \normalsize
    \caption{A general \mapf problem definition.} \label{fig:mapf}
\end{figure}

Using these concepts, we introduce a \mapf problem definition, as in Figure~\ref{fig:mapf}. A \mapf instance is characterized by the quintuple $\seq{A,G,O,cost,\tau}$, and its solution (called a \mapf plan) with a collection $\travcoll$ of traversals.

\paragraph{Remarks: Appearances / disappearances.} This \mapf\ definition is more general than the earlier definitions, as it allows us to explicitly state our assumptions about the appearances / disappearances of agents at their goal / initial locations. 

For instance, consider the following two possible behaviours of agents once they reach their goals:  1) every agent waits at its goal until the traversals of all agents are completed (as in our earlier studies), or 2) every agent disappears from the environment (as in \cite{svancara2019}).  Both cases can be covered with this \mapf\ definition. Suppose that $\ii{reach}_i$ ($x_i \leqs \ii{reach}_i \leqs y_i$) denotes the time step at which an agent $a_i$ reaches its goal. If we assume that the agent $a_i$ disappears when it reaches its goal, then $y_i\eqs \ii{reach}_i$. If we assume that the agent $a_i$ waits at goal until all traversals end, then $\ii{reach}_{i}$ is the time step where the agent reaches its goal and stays there. 

Similarly, consider the following two different behaviours of agents at their initial location: 1) every agent appears at their initial locations, at the time step the agent joins the team (as in our earlier studies), or 2) the agents that are joining the team are allowed to wait outside the environment until their initial locations are unoccupied (as in \cite{svancara2019}). In the former case, for agent $a_i$, the starting time $x_i$ of its traversal is the same as its joining time $join_{i}$.
In the latter case, for agent $a_i$, the joining time $join_i$ is the time at which the agent joins the team and the starting time $x_i$ of its traversal is the time at which the agent enters the environment.

%

\paragraph{Remarks: Cost functions.}
Furthermore, this definition of \mapf also allows different cost functions depending on the needs of the particular application.  

For an agent~$a_i$ with traversal $\trav{i}$ and for a time step $t$, let us first define the {\em cost of waiting} ($cost^i_w(t)$), and the {\em cost of moving} to another vertex ($cost^i_m(t)$):

\vspace{3mm}
\resizebox{0.85\textwidth}{!}{\begin{tabular}{ll}
{$\bullet\ cost^i_w(t) \!\eqs\! \begin{cases}
    1 & \hspace{-1ex} \text{if } f_i(t) \!\eqs\! f_i(t\pluss 1) \text{ for } x_i \leqs t\lts y_i \\
    0 & \hspace{-1ex}\text{otherwise}
\end{cases}$}
&
{\hspace{1.5mm} $\bullet\ cost^i_m(t) \!\eqs\! \begin{cases}
    1 & \hspace{-1ex}\text{if } f_i(t) \neqs f_i(t\pluss 1) \text{ for } x_i \leqs t\lts y_i \\
    0 & \hspace{-1ex}\text{otherwise}
\end{cases}$} 
\end{tabular}}
\vspace{3mm}

\noindent Then we can define the {\em length of its traversal} and the {\em cost of its traversal} $\trav{i}$ (considering the completion of the task, not the waiting afterwards):

\vspace{3mm}
\resizebox{0.9\textwidth}{!}{\begin{tabular}{ll}
$\bullet\ cost^i_{L}(r_{a_i}^{P_i}) \eqs \sum_{t\eqs x_i}^{y_i} cost^i_w(t) \pluss cost^i_m(t)$
&
\hspace{6mm} $\bullet\ cost^i_{T}(r_{a_i}^{P_i}) \eqs \sum_{t\eqs x_i}^{\ii{reach}_{i}} cost^i_w(t) \pluss cost^i_m(t)$
\end{tabular}}
\vspace{3mm}

\noindent 
%
When the distance traveled by an agent $a_i$ is more important than the time spent by the agent, we can consider the {\em cost of its path} $P_i$:
$cost^i_{P}(r_{a_i}^{P_i}) \eqs \sum_{t\eqs x_i}^{y_i} cost^i_m(t) \eqs |P_i|$. 

\smallskip
Now the {\em sum of costs} of a \mapf plan $\mathbf{r}_A^\mathbf{P}$ (i.e., the total time spent by all of the agents until they reach their goals), and the {\em makespan} of a \mapf plan $\mathbf{r}_A^\mathbf{P}$ (i.e., the time step where all agents in $A$ reach their goals) are defined as follows:

\vspace{3mm}
\resizebox{0.9\textwidth}{!}{\begin{tabular}{ll}
$\bullet\ 
cost_{SOC}(\mathbf{r}_A^\mathbf{P}) \eqs \sum_{a_i\ins A} cost^i_T(r_{a_i}^{P_i})$
&
\hspace{6mm} $\bullet\  
cost_{M}(\mathbf{r}_A^\mathbf{P}) \eqs max(cost^i_T(r_{a_i}^{P_i})).$
\end{tabular}}
\vspace{3mm}

\noindent The total distance $cost_{SOP}(\mathbf{r}_A^\mathbf{P})$ traveled by the agents (i.e., the {\em sum of path lengths} of a \mapf plan $\mathbf{r}_A^\mathbf{P})$ is defined similarly, by adding up $cost^i_P$. 

For more details, please see \ref{app:cost}.

\color{black}

\section{\dmapf: Problem Definition}


Consider a given \mapf instance $\seq{A,G,O,cost,\tau}$, and its solution (a \mapf plan) $\mathbf{r}_A^\mathbf{P}$. \dmapf considers changes in the environment or in the team while such a given \mapf plan is being executed. We describe these changes by means of events, defined as follows.

An {\em event} $e$ at a time $t$ ($t\geqs 0$) is characterized by a tuple $\event{t}$, where
\begin{itemize}
    \item $A_t\uparrs$ (resp. $A_t\downarrs$) denotes the set of agents leaving (resp. joining) at time $t$,
    \item $O_t\uparrs$ (resp. $O_t\downarrs$) denotes the set of obstacles removed (resp. added) at time $t$.
\end{itemize}

A sequence $C \eqs\seq{e_0, e_1, \dots, e_m}$ of events leads to changes if, for every $e_k\ins C$, at least one of the sets $A_t\uparrs, A_t\downarrs, O_t\uparrs, O_t\downarrs$ is nonempty.

Given a sequence $C \eqs\seq{e_0, e_1, \dots, e_m}$ of events, the set $A^C_t$ of agents who are present in the environment at time $t$ is defined as follows:
\[A^C_t \eqs \begin{cases}
    A                                                       & \text{if } t\eqs 0, \text{ and } e_0 
    \text{ occurs at some time } z\gts 0  \\
    (A \setminuss A_t\uparrs) \cups A_t\downarrs            & \text{if } t\eqs 0, \text{ and } e_0 
    \text{ occurs at time } 0  \\
    A^C_{t{-}1}                                               & \text{if }
    t\gts 0, \text{ and no event } e_i (i\gts 0) \text{ in } C \text{ occurs at time } t \\
    (A^C_{t{-}1} \setminuss A_t\uparrs) \cups A_t\downarrs    & \text{if }
    t\gts 0, \text{ and some event } e_i (i\gts 0) \text{ in } C \text{ occurs at time } t\\
\end{cases}\]

Given a sequence of events $C \eqs\seq{e_0, e_1, \dots, e_m}$, the  set $O^C_t$ of vertices covered by the obstacles present in the environment at time $t$ is defined in a similar way.

%
%
%

Let us denote by $A^C_{\leqs t}$ the set of agents that were present in the environment before or at some time $t$, as a sequence $C$ of events take place. 

We say that a sequence $C \eqs\seq{e_0,e_1,\dots,e_m}$ of events is {\em valid} if the following hold:

           \vspace{2mm}
\noindent (i) for the event $e_0 \eqs \event{t}$,
    \begin{itemize}
        \item $A_t\uparrs \subseteqs A$, 
            (i.e., leaving agents are present in the environment initially),
        \item $A_t\downarrs \caps A \eqs \emptyset$ (i.e., joining agents are not already in the environment),
        \item $O_t\uparrs \subseteqs O$,
        (i.e., removed obstacles are in the environment initially),
        \item $O_t\downarrs \caps (O \setminuss O_t\uparrs) \eqs \emptyset$ (i.e., new obstacles are different from the obstacles that are already in the environment), and
    \end{itemize}

\noindent (ii) for every event $e_k \eqs \event{t}$ ($0 \lts k\leqs m$),
    \begin{itemize}
        \item $A_t\uparrs \subseteqs A^C_{t{-}1}$ (i.e.,  agents leaving at time $t$, are present there at time $t{-}1$),
        \item $A_t\downarrs \caps A^C_{\leqs t{-}1} \eqs \emptyset$ (i.e., agents joining at time $t$ have never been in the environment previously (with the same id)),
        \item $O_t\uparrs \subseteqs O^C_{t{-}1}$ (i.e., obstacles removed at time $t$ are in the environment at time $t{-}1$),
        \item $O_t\downarrs \caps (O^C_{t{-}1} \setminuss O_t\uparrs) \eqs \emptyset$ (i.e., new obstacles added at time $t$ are not already present in the environment), and
    \end{itemize}

\noindent (iii) for every two events $e_k\eqs\event{t}$ and $e_{k\pluss 1}\eqs\event{z}$ ($0 \leqs k\lts m$),
    \begin{itemize}
        \item
        $t \lts z$ (i.e., event $e_k$ occurs before event $e_{k\pluss 1}$).
    \end{itemize}

Note that, for every event $e_k \eqs \event{t}$ in a valid event sequence $C$, 1)~for every agent $a_i \ins A_t\downarrs $, $t\eqs join_i$, and 2)~for every agent $a_j \ins A_t\uparrs $, $t \eqs y_j$.

\begin{figure}[t]
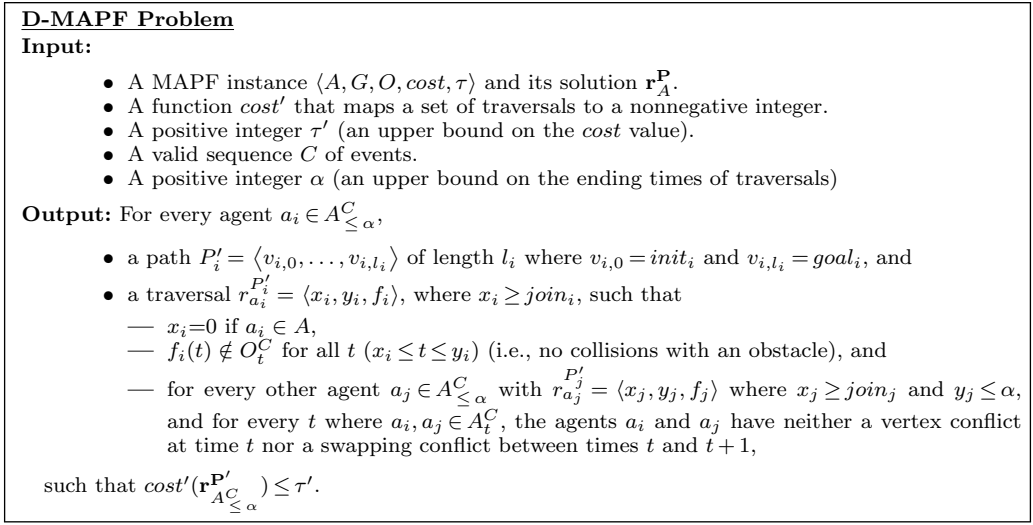

    \centering%
    \fbox{
    \begin{minipage}{0.98\columnwidth} 
        \footnotesize
        {\bf\underline{\dmapf Problem}}
        \vspace{.25ex}

        {\bf Input:}
          \begin{itemize}[leftmargin=14mm]
            \item A \mapf instance $\seq{A,G,O,cost,\tau}$ and its solution $\mathbf{r}_A^{\mathbf{P}}$.
            \item A function $cost'$ that maps a set of traversals to a nonnegative integer.
            \item A positive integer $\tau'$ (an upper bound on the $cost$ value).
            \item A valid sequence $C$ of events.
            \item A positive integer $\alpha$ (an upper bound on the ending times of traversals)
        \end{itemize} 
        {\bf Output:} For every agent $a_i \ins A^C_{\leqs {\alpha}}$, 
        \begin{itemize}[leftmargin=14mm]
            \item a path $P'_i \eqs \seq{v_{i,0}, \dots, v_{i,l_i}}$ of length $l_i$ where $v_{i,0}\eqs init_{i}$ and $v_{i,l_i}\eqs goal_{i}$, and 
            \item a traversal $\travp{i}$, where $x_i \geqs join_{i}$, such that 
            \begin{itemize}
                \item $x_i{=}0$ if $a_i\in A$,
                \item $f_i(t) \notin O^C_t$ for all $t$ ($x_i \leqs t \leqs y_i$) (i.e., no collisions with an obstacle), and 
                \item for every other agent $a_j\ins A^C_{\leqs {\alpha}}$  with $\travp{j}$ where $x_j \geqs join_j$ and $y_j\leqs \alpha$, and for every $t$ where $a_i, a_j \ins A^C_t$, the agents $a_i$ and $a_j$ have neither a vertex conflict at time $t$ nor a swapping conflict between times $t$ and $t\pluss 1$,
                \end{itemize}
            \end{itemize} 
       \hspace{2mm} such that $cost'(\mathbf{r}_{A^C_{\leqs \alpha}}^\mathbf{P'}) \leqs \tau'$. 


    \end{minipage}} \normalsize
    \caption{A general definition for \dmapf problem.} \label{fig:dmapf}
\end{figure}

Based on the concepts and notation defined above, we introduce a general definition for  the \dmapf problem, as in Figure~\ref{fig:dmapf}.

%

\paragraph{Remarks.} Our \dmapf\ definition can be easily extended to include different constraints, as needed by applications. For instance, during the applications of our studies at warehouses with mobile robots, we have observed the need for another constraint that prevents conflicts due to robots following each other too closely.
For those studies, we have defined {\em following conflicts} as follows, and extended the \dmapf\ definition accordingly:
For two agents $a_i,\,a_j$ with traversals $\trav{i}$ and $\trav{j}$, respectively, a {\em following conflict} 
    occurs between $a_i$ and $a_j$ between times $t$ and $t\pluss 1$, where $\max(x_i,x_j)\leqs t \lts \min(y_i,y_j)$, if $f_i(t) \eqs f_j(t\pluss 1)$. 
\section{A General Framework for Solving \dmapf Problems}

\begin{figure}[t]
    \centering
    \includegraphics[scale = 0.4]{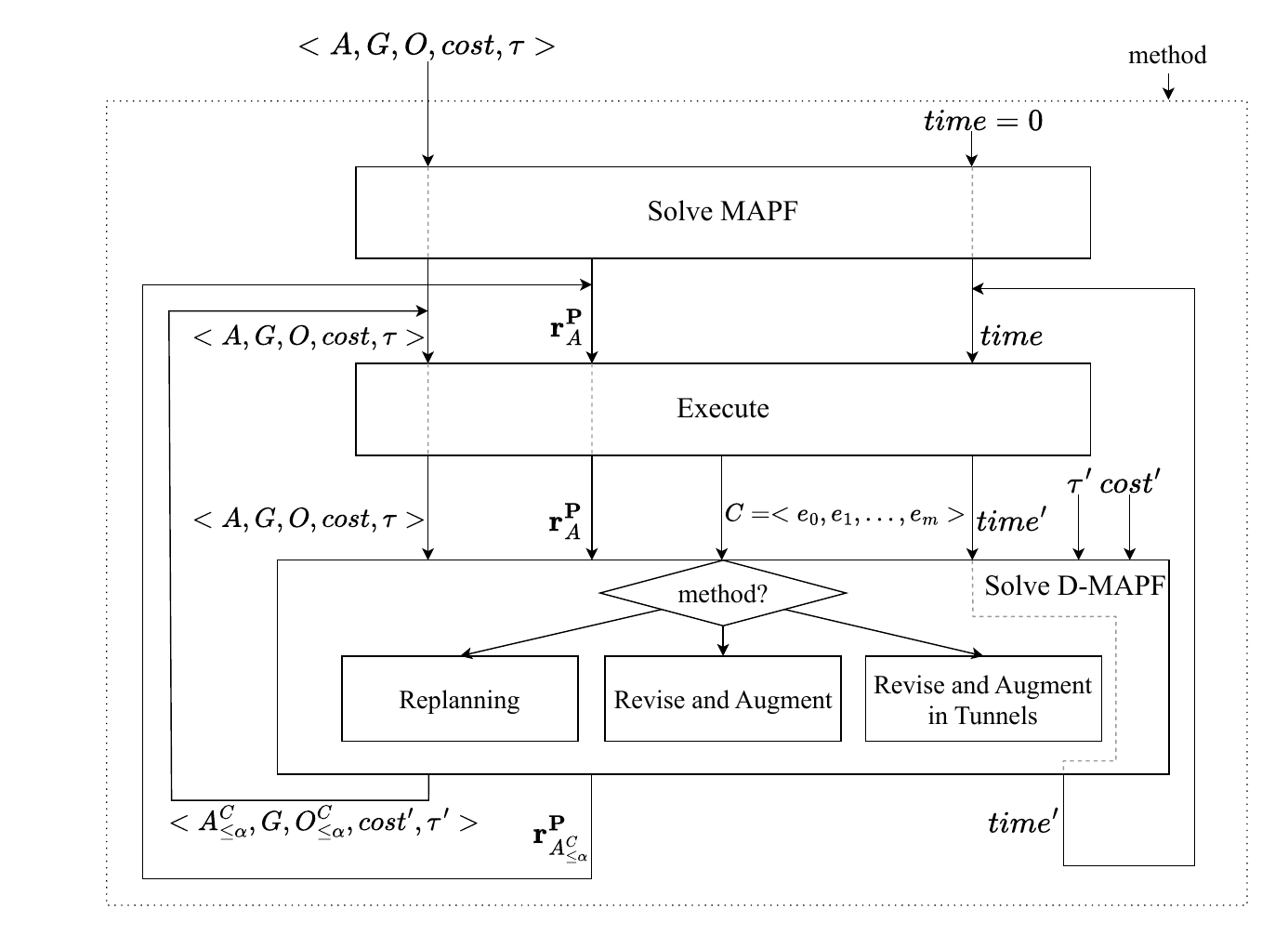}
    \caption{Overall architecture of a general framework for \dmapf.}
    \label{fig:dmapf_algo} 
\end{figure}

\underline{Overall architecture.} We introduce a general framework for \dmapf problems, based on multi-shot computation, and utilizing one of specified methods to solve \dmapf.
Figure~\ref{fig:dmapf_algo} shows an overall architecture of this framework, considering online execution of the computed plans.

In an online execution, the changes that will occur in the environment are not known before the start of the execution. To detect the changes, we consider the environment being monitored by a central agent. When this central agent detects a change in the environment during the execution of a computed plan, a new plan is computed and the agents start executing the new plan.

Our algorithm takes a \mapf instance $\seq{A,G,O,cost,\tau}$ and a method to solve \dmapf as input. 
\begin{enumerate}
    \item Initially, a \mapf solution is computed for this instance, starting from time $0$, shown with the ``Solve \mapf'' block.

\item  Then, the computed solution $\travcoll$ is executed step by step until a change is detected by the central agent at time $time'$. This execution process is shown with the ``Execute'' block.

\item The instance $\seq{A,G,O,cost,\tau}$ and its solution $\travcoll$ being executed are the inputs for the next step, ``Solve \dmapf'', as well as a sequence of events $C \eqs \seq{e_0,e_1,\dots,e_m}$, the cost function $cost'$ and its upper bound $\tau'$. In this part, \dmapf problem is solved using the method specified at the beginning of the algorithm.

 \item Once a \dmapf solution $\mathbf{r}^\mathbf{P}_{A^C_{\leqs\alpha}}$ is found and the current instance is updated as $<A^C_{\leqs\alpha}, G, O^C_{\leqs\alpha}, cost', \tau'>$, the execution continues from time step $time'$ with the ``Execute'' block. The execution continues until the end of the computed plan.

\end{enumerate}

\underline{Multi-shot solving.}
For the solving processes ``Solve \mapf'' and ``Solve \dmapf'' blocks, we utilize multi-shot ASP~\citep{gebser2019multi}.

Multi-shot solving aims to handle continuously changing logic programs. A multi-shot ASP program is able to grow and be updated with the changing knowledge to solve a problem. A multi-shot ASP program considers a program splitted into multiple parts. The inclusion of these subprograms into the solving process is maintained by a controller program from the outside. The subprograms generally serve different purposes and can be grouped as: static parts that are grounded once and not changed throughout the program (usually called the \verb|base| program), cumulative parts that can be added multiple times with different parameter values, such as time step in dynamic domains (usually called the \verb|step| program), and volatile parts that are added for a step and removed in the next step, checking whether the stopping condition is satisfied (usually called the \verb|check| program). Enabling and disabling rules in the \verb|check| program at different steps is done with \verb|external| atoms, through the outside controller program.

In our algorithm for solving \dmapf problem, the \verb|base| program contains the \mapf instance and the initial step of the traversal. The \verb|step| program has a parameter \verb|t|, denoting the time step of the plan, and it generates the plan recursively for time step \verb|t| and adds the collision constraints for time \verb|t|. The \verb|check| program also has a parameter \verb|t| and an external atom called \verb|query| for enabling/disabling the rules in this program. This program verifies whether every agent reached their goals at time~\verb|t|. 
These programs are depicted in the left side of~Figure~\ref{fig:programs}, and explained more in detail in \ref{app:asp_formulations}.

\begin{figure}[t] 
\begin{minipage}{0.48\textwidth}
\begin{verbbox}[\mbox{}]
#program base.
time(0).
plan(A,0,X):- init(A,X), agent(A).
  
#program step(t).
time(t).  
{plan(A,t,X)}1 :- plan(A,t-1,X), agent(A).
{plan(A,t,Y): edge(X,Y)}1 :- plan(A,t-1,X), agent(A).
:- {plan(A,t,X): vertex(X)}0, agent(A).
:- 2{plan(A,t,X): vertex(X)}, agent(A).

:- 2{plan(A,t,X): agent(A)}, vertex(X), time(t).
plan_to(X,Y,t,0) :- plan(A,t-1,X), plan(A,t,Y), 
  edge(X,Y), agent(A).
:- plan_to(X,Y,t,_), plan_to(Y,X,t,_).

goal_reached(A,t) :- goal(A,X), plan(A,t,X), agent(A). 
goal_reached(A,t) :- goal_reached(A,t-1).

#program check(t).
#external query(t).
:- not goal_reached(A,t), agent(A), query(t).

:- not goal(A,X), plan(A,t,X), agent(A), query(t).
\end{verbbox}
\resizebox{\textwidth}{!}{\fbox{\theverbbox}}
\end{minipage}
\hspace*{2mm}
\begin{minipage}{0.48\textwidth}
\begin{verbbox}[\mbox{}]
#program newAgent(a,i,g,k).
agent(a).
init(a,i). 
goal(a,g). 

plan(a,k,X) :- init(a,X), agent(a).
{plan(a,T,X)}1 :- plan(a,T-1,X), agent(a), time(T), T >= k.
{plan(a,T,Y): edge(X,Y)}1 :- plan(a,T-1,X), agent(a), 
  time(T), T >= k.
:- {plan(a,T,X): vertex(X)}0, agent(a), time(T), T >= k.
:- 2{plan(a,T,X): vertex(X)}, agent(a), time(T), T >= k.

:- 2{plan(A,T,X): agent(A)}, vertex(X), time(T), T >= k.
plan_to(X,Y,T,k) :- plan(a,T-1,X), plan(a,T,Y), 
  edge(X,Y), agent(a), time(T), T>=k.
:- plan_to(X,Y,T,_), plan_to(Y,X,T,_), time(T), T>=k.

goal_reached(a,T) :- goal(a,X), plan(a,T,X), 
  agent(a), time(T), T>=k.
goal_reached(a,T) :- goal_reached(a,T-1), time(T), T>=k.

:- not goal_reached(a,T), agent(a), query(T).
:- not goal(a,X), plan(a,T,X), agent(a), query(T).
\end{verbbox}
\resizebox{\textwidth}{!}{\fbox{\theverbbox}}
\end{minipage} %
\caption{{The multi-shot ASP programs used for solving \mapf and \dmapf.}}
\label{fig:programs} \color{black}
\end{figure} 

\underline{Solving \mapf, using multi-shot ASP.}
``Solve \mapf'' procedure utilizes the programs and solves the problem following these steps: 
\begin{enumerate}
    \item The controller program starts with time step \verb|t=0|, grounds the \verb|base| program and \verb|check(0)| program, assigns the external atom \verb|query(0)| and solves the current program, if the solver does not return a value (SAT or UNSAT), continues with step 2. Otherwise the computation ends. 
    \item Time step value \verb|t| is increased by 1, the external atom \verb|query(t-1)| is released and \verb|step(t)| and \verb|check(t)| are grounded, the external atom \verb|query(t)| is assigned and the current program is solved.
    \item If the solver returns a value or the makespan limit is reached, solving ends and the solution is returned, otherwise step 2 is repeated.
\end{enumerate}

The solution found in ``Solve \mapf'' is passed to the ``Execute'' block and executed step by step until a change is detected by the central agent and \dmapf is solved. The visualization of these steps of solving can be found in Figure~\ref{fig:solve_fnc} in \ref{app:asp_formulations}.

\underline{Solving \dmapf, using multi-shot ASP.}
``Solve \dmapf'' has a similar procedure as described above for ``Solve \mapf'', but has additional parts for maintaining the new agents. \dmapf solving uses the existing ground program of \mapf, with the same controller program. There is an additional program, called \verb|newAgent|, for each of the three methods, presented on the right side of Figure~\ref{fig:programs}. This program has 4 parameters: the agent \verb|a|, its initial position \verb|i|, its goal position \verb|g| and the starting time of its traversal~\verb|k|. If multiple agents are added at the same time step, then there will be a different \verb|newAgent| program for each agent. 

\section{A New Method for \dmapf: Revise-and-Augment-in-Tunnels}

In this study, we consider the following two existing approaches,  to solve \dmapf\ problems.   The \textit{Replan-All} method (as in~\cite{svancara2019}) discards the existing \mapf plan and re-solves \mapf for all agents considering the changes in the team. 
The \textit{Revise-and-Augment} method (as in~\cite{bogatarkanPE19}), on the other hand, reuses the existing plan: when a change occurs, the plans of existing agents are revised by rescheduling their waiting times, while plans are computed for the newly joining agents. In this approach, while revising plans, the order of vertices that the agent visits in the \mapf solution is preserved in the \dmapf solution for every existing agent. 

%
%


We also introduce a new approach (called 
{\it Revise-and-Augment-in-Tunnels}) to solve \dmapf problems, that combines the advantages of Replan-All and Revise-and-Augment methods. Revise-and-Augment-in-Tunnels aims to reuse the previously computed plans in the spirit of Revise-and-Augment, but in a more relaxed way in the spirit of Replan-All. While revising plans, instead of requiring that every existing agent follow their existing paths only, Revise-and-Augment-in-Tunnels allows each agent to visit some other vertices neighboring their path. Note that, unlike Revise-and-Augment, this method allows the agents to visit the vertices of their paths in a different order. 

\begin{figure}[h!]
    \centering
    \resizebox{.8\columnwidth}{!}{\begin{tabular}{ccc}
         \includegraphics[width=0.25\textwidth]{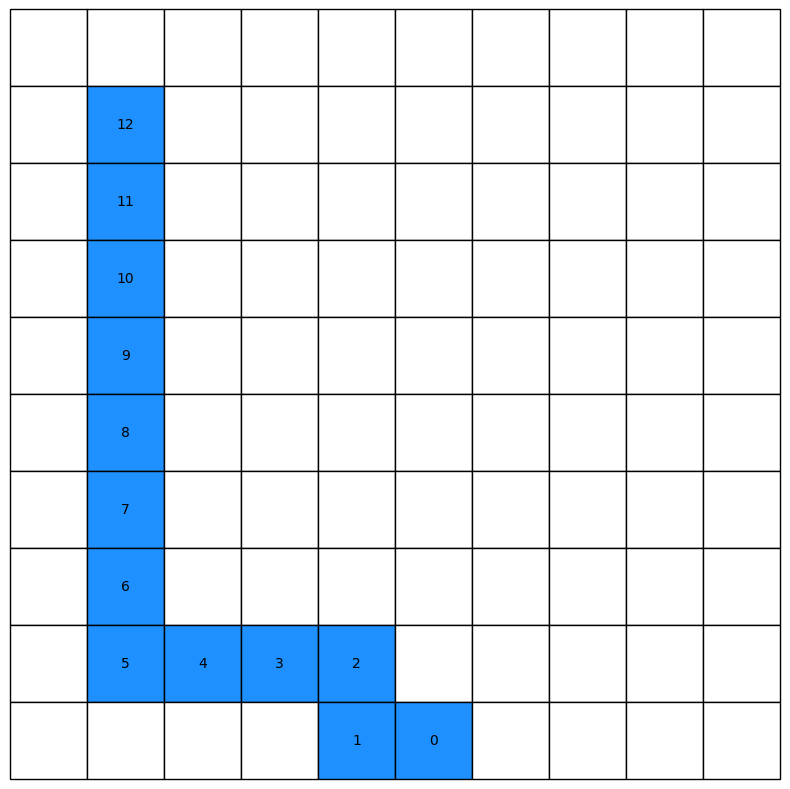} &
         \includegraphics[width=0.25\textwidth]{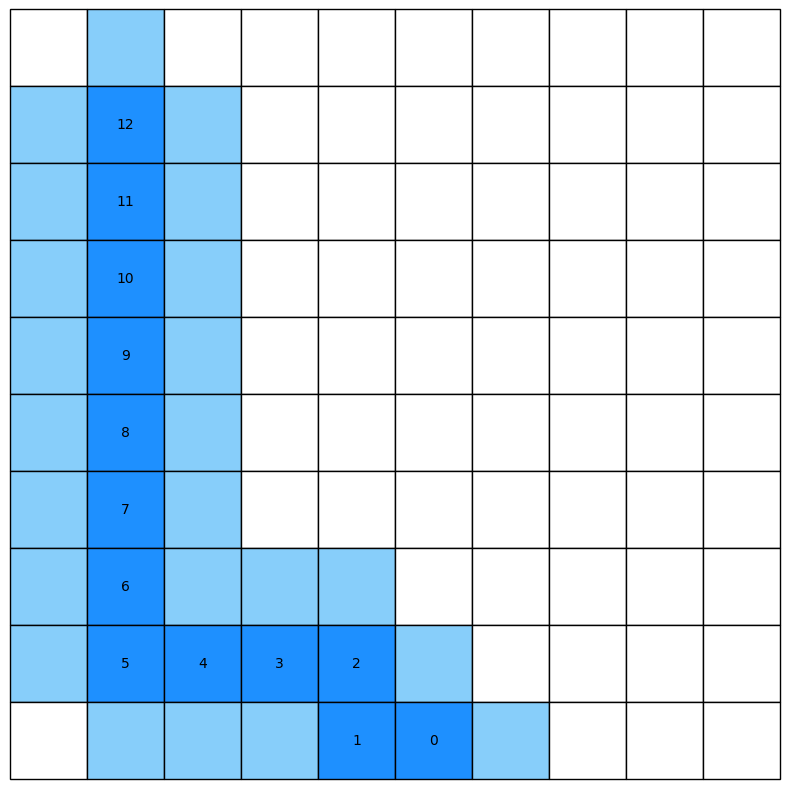} &
         \includegraphics[width=0.25\textwidth]{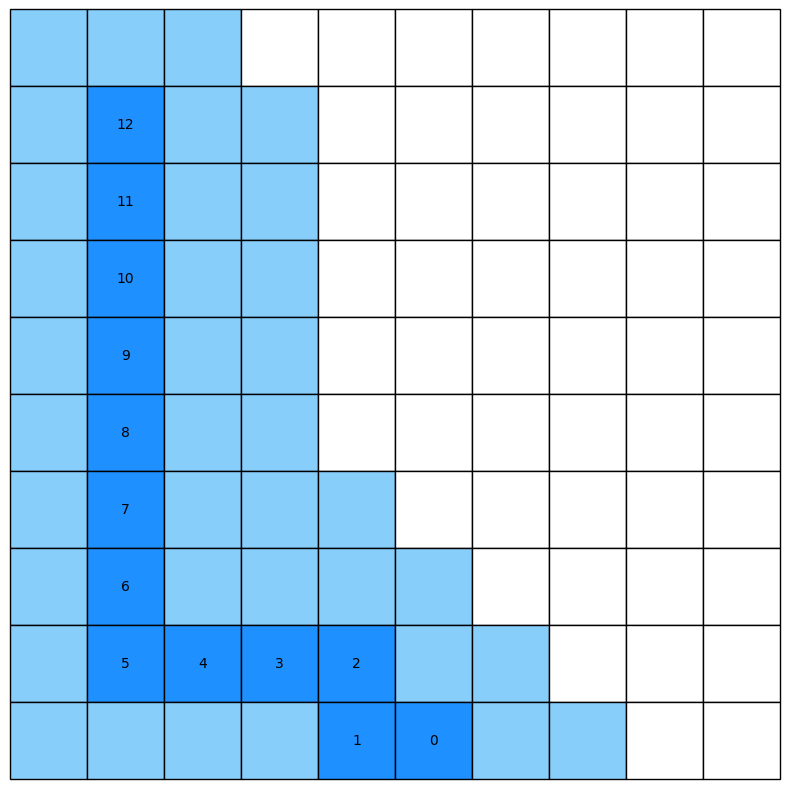} 
    \end{tabular}}
    \caption{Tunnels with widths 0 (left), 1 (middle) and 2 (right).}
       \label{fig:tunnel_idea}     
\end{figure}

Intuitively, each existing agent is allowed to follow some path in a “tunnel” of a specified “width”. Such a tunnel consists of the vertices included in the agent’s existing path, the neighboring vertices within a Manhattan distance of ``width'' from the path, and the edges of the graph that connect these vertices. Figure~\ref{fig:tunnel_idea} shows sample tunnels with different width values for a sample agent. Note that a tunnel with a zero width contains only the path of the agent and the edges between them. 

%

%
 
Formally, the {\em tunnel} $T^w_i$ of an agent $a_i$ with respect to its path $P_i$ in $G$, with width $w\geqs 0$, is the induced subgraph of the set $\{u: v\ins P_i,\ u\in V,\ 0 \leqs d_M(u,v) \leqs w\}$  of vertices, where $d_M$ denotes the Manhattan distance.

    
For every agent $a_i \ins A$ with a path $P_i$, Revise-and-Augment-in-Tunnels considers the following {\em tunnel constraints} while revising plans: 
    every vertex $v_{i,j}$ visited by $a_i$ in the revised plan should be in $T^w_i$, and, for every $\seq{v_{i,j},v_{i,j+1}}$ followed by $a_i$ in the revised plan, there exists an edge  $\{v_{i,j},v_{i,j+1}\}$ in $T^w_i$.
    
Note that, essentially, Revise-and-Augment-in-Tunnels solves \dmapf\ extended with such tunnel constraints.

\paragraph{Remarks.} 
To better understand the difference between Revise-and-Augment and Revise-and-Augment-in-Tunnels with tunnel width $0$, consider the following example: An agent $a_i$ has a path $P_i \eqs \seq{A,B,C,D}$ and its traversal $\travi{i}$ with $f_i \eqs \seq{A,A,B,C,C,D}$. At time $t \eqs 1 $ new agents join the environment and the traversal for $a_i$ will be revised using Revise-and-Augment-in-Tunnels with tunnel width $0$. Agent $a_i$ can have a possible revised traversal $r_{a_i}^{P'_{i}}$ with $f_i\eqs\seq{A,A,B,C,B,C,D}$ (if the edges are undirected). This revised traversal is not possible to obtain with the Revise-and-Augment method, since the order of vertices in the path are not allowed to change. 



\section{Three Multi-Shot ASP Methods for \dmapf: \\ Replan-All, Revise-and-Augment, and Revise-and-Augment-in-Tunnels}

\underline{Replan-All, using multi-shot ASP.} If the method used for solving \dmapf is replanning, then only the \verb|newAgent| program and the programs from ``Solve \mapf'' are used with the following steps, assuming we have an existing ground program until some makespan~\verb|m|, and the time of change \verb|k|: 
\begin{enumerate}
    \item For every agent \verb|a| added at time \verb|k|, a \verb|newAgent| program for the agent \verb|a|, its initial and goal locations \verb|i| and \verb|g| and the starting time \verb|k| is grounded. This program recursively computes a plan for the agent starting from time \verb|k| until time \verb|m| and adds the collision constraints for the agent and the goal condition for the agent.
    \item Once all the \verb|newAgent| programs are grounded for time \verb|k|, the cumulative ground program containing old and new agents is solved.
    \item If the solver returns a solution, it is passed to the execution. Otherwise, the algorithm tries to find a solution with a makespan \verb|m+1|. This is done with the same steps in ``Solve \mapf''. However, instead of starting from \verb|t=0|, the procedure starts from \verb|t=m| since the time steps from 0 to \verb|m| are already grounded for the old and new agents. Therefore, only steps 2 and 3 in ``Solve \mapf'' are used.
\end{enumerate}

Since the program does not have any knowledge about the already executed part of the plan, we need make sure that the existing agents' location at time \verb|k| is the same in the new plan, to avoid them jumping to a disconnected location in the next step. For this purpose, we use the controller program and utilize assumptions of multi-shot {\em clingo}. We set the truth value of each plan atom in the already executed part of the existing solution to true, making sure that those atoms are included in the answer set in the new solution. This is done after every change in the environment, before starting the procedure of solving \dmapf.

\begin{figure}[t] 
    \begin{minipage}{0.8\textwidth}
\begin{verbbox}[\mbox{}]
#program path_constraints(t,k).
#external path_query(t,k). 

plan_pair(A,X,Y,t,k) :- plan(A,T-1,X), plan(A,T,Y), X!=Y, old_agent(A), path_query(t,k). 
plan_pair_count(A,X,Y,C1,t,k) :- C1 = #count{X,Y,T: plan(A,T-1,X), plan(A,T,Y), X!=Y},
  plan_pair(A,X,Y,t,k), old_agent(A), path_query(t,k).

:- plan_pair_count(A,X,Y,C1,t,k), path_pair_count(A,X,Y,C), C1!=C, old_agent(A), path_query(t,k).

:- plan(A,_,X), not path(A,X,_), old_agent(A), path_query(t,k). 
:- not plan(A,_,X), path(A,X,_), old_agent(A), path_query(t,k). 
\end{verbbox}
\resizebox{\textwidth}{!}{\fbox{\theverbbox}}
\end{minipage} 
\caption{{The multi-shot ASP program used for solving \dmapf with Revise-and-Augment.}}
\label{fig:pathconstraints} \color{black}
\end{figure}

\underline{Revise-and-Augment, using multi-shot ASP.}
When solving \dmapf with Revise-and-Augment, in addition to the \verb|newAgent| program and the programs from ``Solve \mapf'', we have an additional program called \verb|path_constraints| (shown in Figure~\ref{fig:pathconstraints}, and explained more in detail in \ref{app:asp_formulations}). It has two parameters, one of them is the time step \verb|t| that the constraints are being grounded for, and the other one is \verb|k|, the time step of the last change. This program contains an external atom, called \verb|path_query(t,k)| for adding and removing the constraints at each step. When a change is observed during the execution, in addition to the assumptions explained above, we add additional atoms to the {\em clingo} control, to be used with the \verb|path_constraints| program. These atoms add the knowledge about the order of atoms in the existing paths for the agents and which agents are the relevant ones for preserving the paths. Detailed explanation about these atoms and how they are used in the program can be found in \ref{app:asp_formulations}. Once the relevant knowledge is added, the following steps are used for computing the revised and augmented solution: 
\begin{enumerate}
    \item \verb|newAgent| programs are grounded same as the step 1 of Replan-All.
    \item If there is a \verb|path_query| assigned from earlier, it is released. The \verb|path_constraints| program  with parameters \verb|m| and \verb|k| is grounded and its \verb|path_query| atom is assigned with the same parameters.
    \item The cumulative program, with all agents and  relevant path constraints, is solved.
    \item If the solver returns a solution, it is passed to the execution. Otherwise, the algorithm tries to find a solution with a makespan \verb|m+1|. This is done similarly with ``Solve \mapf'' but has additional steps for preserving the paths.
    \begin{enumerate}
        \item Similar to Replan-All, solving procedure starts with \verb|t=m| from step 2 of ``Solve \mapf''. Then \verb|t| value is increased by 1, the external atom \verb|query(t-1)| is released, \verb|step(t)| and \verb|check(t)| are grounded, and 
        \verb|query(t)| is assigned.
        \item 
        Previously used \verb|path_query| is released, the \verb|path_constraints(t,k)| program is grounded and \verb|path_query(t,k)| is assigned. The program containing current step and the constraints for path is solved.
        \item If the solver returns a value or the makespan limit is reached, solving ends and the solution is returned, otherwise steps (a) and (b) are repeated.
    \end{enumerate}
\end{enumerate}

\underline{Revise-and-Augment-in-Tunnels, using multi-shot ASP.}
To solve \dmapf with this method utilizing multi-shot ASP, one of the following two approaches can be considered

In the first approach, tunnels are constructed at the plan generation part. In \verb|step| and \verb|newAgent| programs, we update the choice rules for moving to an adjacent vertex: 

\verb|{plan(A,t,Y): edge(X,Y), tunnel(A,Y)}1 :- plan(A,t-1,X), agent(A).| 

\begin{enumerate}
\item
Initially, when grounding the \verb|base| program for \mapf, all vertices in the environment are added as part of the tunnel of the agents, with external \verb|tunnel(A,X)| atoms from the controller program, meaning that vertex~\verb|X| is in the tunnel of agent~\verb|A|. Then, the controller follows the same steps from ``Solve \mapf''.

\item
Once a change in the environment occurs, the paths of the existing agents are extracted from the solution, and the tunnels are computed according to the width value. The tunnel of an existing agent is only computed once, at the first time step when it becomes an old agent after a change in the environment. The external \verb|tunnel| atoms that represent the locations 
outside of the tunnel of an agent are released. This disables the plan generation rules containing the vertices outside of the tunnel of the agent.

\item
After this point, solving process continues the same way with replanning, and, after every change, the tunnels of the agents that become old agents in that time step are updated before moving on with the \verb|newAgent| program.
\end{enumerate}

In the second approach, the locations that are outside an agent's tunnel are considered as forbidden locations for that agent. Accordingly, constraints are added to ensure that no agent visits their forbidden locations in its plan. The procedure for adding these constraints is similar to changes done for the Revise-and-Augment method. 
\begin{enumerate}
\item
The forbidden locations are extracted from the existing solution and added as \verb|forbidden(A,X)| atoms to the program by the controller similar to the \verb|path_order| atoms in Revise-and-Augment.
\item
After a change in the environment, a program called \verb|forbidden_locations| containing the constraint of not visiting forbidden vertices is added for every existing agent, if it is not already added before for that agent.
\item
The forbidden locations of an agent do not change over time, therefore there are no query atoms in this program and it is not added and removed at each step. 
\end{enumerate}


\section{Experimental Evaluations}

We evaluate the performance and usefulness of our methods by comparing them with each other, in terms of computation time and quality of solutions. We implement our framework using multi-shot ASP with {\em clingo} 5.8.0 and Python 3.8.10. 

\smallskip\noindent{\it \underline{Investigating the computational performance of multi-shot ASP-based \dmapf methods}.}
With the first set of experiments, our goal is to observe how each of these methods perform in terms of computation time, with different number of agents joining the environment at different times. For this purpose, we generated instances with an empty grid of size $20\times20$, initially having 20 or 30 agents. We added 5 or 10 new agents to these instances, at different time steps with different group sizes. A total of 25 instances were generated, with 5 different initial and new agent setups.

In these empty-grid instances, the initial locations and the goal locations of the agents are picked such that each initial-goal pair is at the opposite diagonal corners of each other. This setup results with longer plans in empty grids, allowing us to observe the changes in the computation times more easily, since every step added to the plans has an impact on the computation time.

\begin{figure}[t]
    \centering
    \includegraphics[width=.95\textwidth]{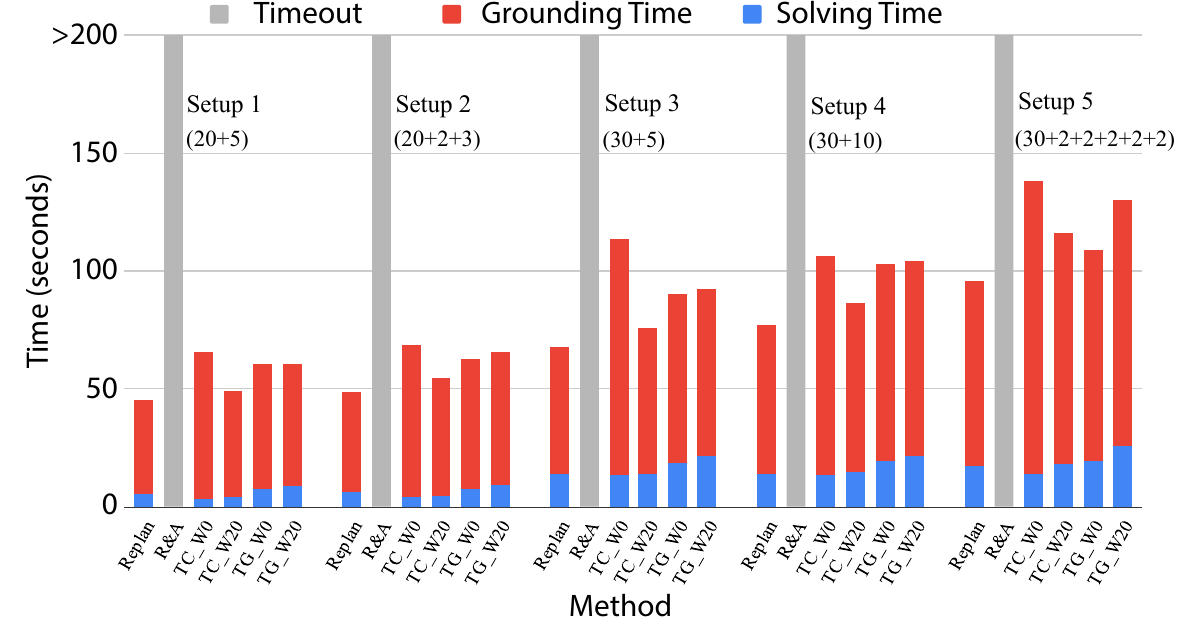}
    \caption{{Average grounding and solving times for 5 setups on an empty $20{\times}20$ grid.} }
    \label{fig:bar_chart_empty_grid}
\end{figure}

{Figure~\ref{fig:bar_chart_empty_grid} shows the total computation times for 5 different setups of these instances with different methods. Each bar shows the average time of 5 instances with the same setup.  For instance, Setup~1~(20+5) has initially 20 agents, and 5 agents join at time~1. Setup~5~(30+2+2+2+2+2) has initially 30 agents; then 2 agents join at time 1, 2 agents join at time 2, 2 agents join at time 3, 2 agents join at time 4, and 2 agents join at time~5.} 

The computation times consist of two parts, grounding and solving time, taken from {\em clingo}. We run Revise-and-Augment-in-Tunnels with tunnel constraints (shown as TC) and Revise-and-Augment-in-Tunnels with tunnels in generate (shown as TG) with two different tunnel width values of 0 and 20. Width 0 only uses the vertices of the path of an agent, bringing the solution closer to Revise-and-Augment (shown as R\&A), and width~20 covers the whole environment, therefore, the same environment as Replan-All. 

Note that in the figure, the computation times greater than 200 seconds are not displayed exactly. 
{A timeout of 200 seconds was set for the experiments and none of the instances was solved with Revise-and-Augment (R\&A) within this time limit.} The detailed results for each instance and method can be found in \ref{app:empty_grid_results}, including the separate times in each stage of solving \dmapf, where finding a new solutions after a change is called a stage. 

%
%
%
%

\smallskip \noindent{\it Observations and discussions about the computational performance of the methods.} 
From these results, we observe that Revise-and-Augment has the highest computation times among all methods and the fastest performing method is Replan-All. Both approaches for solving Revise-and-Augment-in-Tunnels, TC and TG, perform close to each other and Replan-All, but are not as fast as Replan-All for all of our instances.

We observe that, for TC, when the tunnel width increases, the computation times decrease. This decrease is visible mostly in grounding times, whereas the solving times do not differ significantly for different tunnel widths. This shows the effect of grounding the constraints for the forbidden vertices, which are added for each vertex outside an agent's tunnel, for every existing agent. As the tunnel width increases, the number of forbidden vertices decreases, reducing the number of added constraints and therefore the grounding time for constraints. 

On the other hand, for TG, when the tunnel width increases, we observe some increase in computation times. This is due to the higher number of tunnel atoms considered for generating the plans for wider tunnels. As the tunnel width gets larger, the number of tunnel atoms considered for each agent increases accordingly. The number of tunnel atoms directly impacts the number of choice rules being used for generating the new plans, therefore increasing the computation time.  The effect of the tunnel atoms on computation times can clearly be observed in the computation times of initial \mapf solving stage. Initially, every vertex is a part of every agent's tunnel and for our instances this creates an increase around 10 seconds in the total time of solving \mapf compared to other methods. 

If we consider the case where the \mapf instance is the same and the new agents are added at the same time steps but the number of new agents added at those time steps are different  (e.g., Instances 3 and 4), the increase in the computation times with the increase in the number of agents is visible in the results of all methods except R\&A.  For example, for Replan-All the computation time for finding a solution after adding 5 new agents in Instance 3 is 19.01 seconds while the time for finding a solution after adding 10 new agents in Instance 4 is 28.96 seconds. Similar increases can be observed for TC and TG. 

Consider Instance 4 and Instance 5, where the \mapf instance and the total number of new agents are the same, but they are joining as different groups at different times. For all of our methods, the difference of total times are visible in the results. When the same agents are added at different times, the computation takes longer. This is mainly due to the increase in the makespan of the whole \dmapf solution, caused by the agents starting their plans at a later time.

\smallskip\noindent{\it \underline{Investigating the quality of solutions}.}
In our second set of experiments, we used instances with obstacles adapted from \mapf benchmarks in~\cite{SternSFK0WLA0KB19}, as follows: 1)~We selected three types of $32\times32$ grids as environments: \textit{random-32-32-10}  contains random obstacles in the 10\% of the environment, \textit{random32-32-20} has random obstacles in the 20\% of the environment, and \textit{room32-32-4} divides the environment into small rooms and connects them through ``doors''. 2)~We scaled these grids to size $20\times20$ and utilized the random scenarios provided for these grids for creating our instances. 3)~From all random scenarios, we extracted the initial and goal location pairs that do not overlap with obstacles in any of these environments. {4) We merged these pairs to a list, and generated 5 base \mapf instances by randomly selecting 20 agents from the merged list. 
5) Then, for each of these \mapf instances,  we created 10 different \dmapf instances, containing 20 new agents randomly selected from the list, that are not used in its \mapf instance. This resulted in 50 \dmapf instances that can be used in the selected environments, making 150 instances in total. For these experiments, we assume that all new agents are added at $t\eqs0$ before the execution but after computing a \mapf plan.}

We used these instances to investigate the quality of solutions computed with the tunnels. For the old agents, we examined how their initially computed plans change after the new agents are added to the environment. Table~\ref{tab:mapf_benchmarks} shows a sample result, for each environment for the same set of agents. 

\begin{table}[t]
    \caption{Results for the same \dmapf instance in three different 20x20~environments. }
    \label{tab:mapf_benchmarks}
    \centering
     \tablefont\resizebox{.95\columnwidth}{!}{\begin{tabular}{@{\extracolsep{\fill}}cccccccc} 
    \topline 
    \multirow{2}{0.1\linewidth}{Grid} & \mapf & \dmapf & \#Plan  & \#Path  & \multicolumn{3}{c}{\#Diverted Agents} \\
                                      & Mkspn &  Mkspn & Changes & Changes & \multicolumn{3}{c}{[Amount of Divergence in Replan-All]} \\ 

    \hline

\multirow{5}{0.12\linewidth}{
    \begin{minipage}{\linewidth} \vspace{-2mm}
        \centering
        \includegraphics[width=0.6\linewidth]{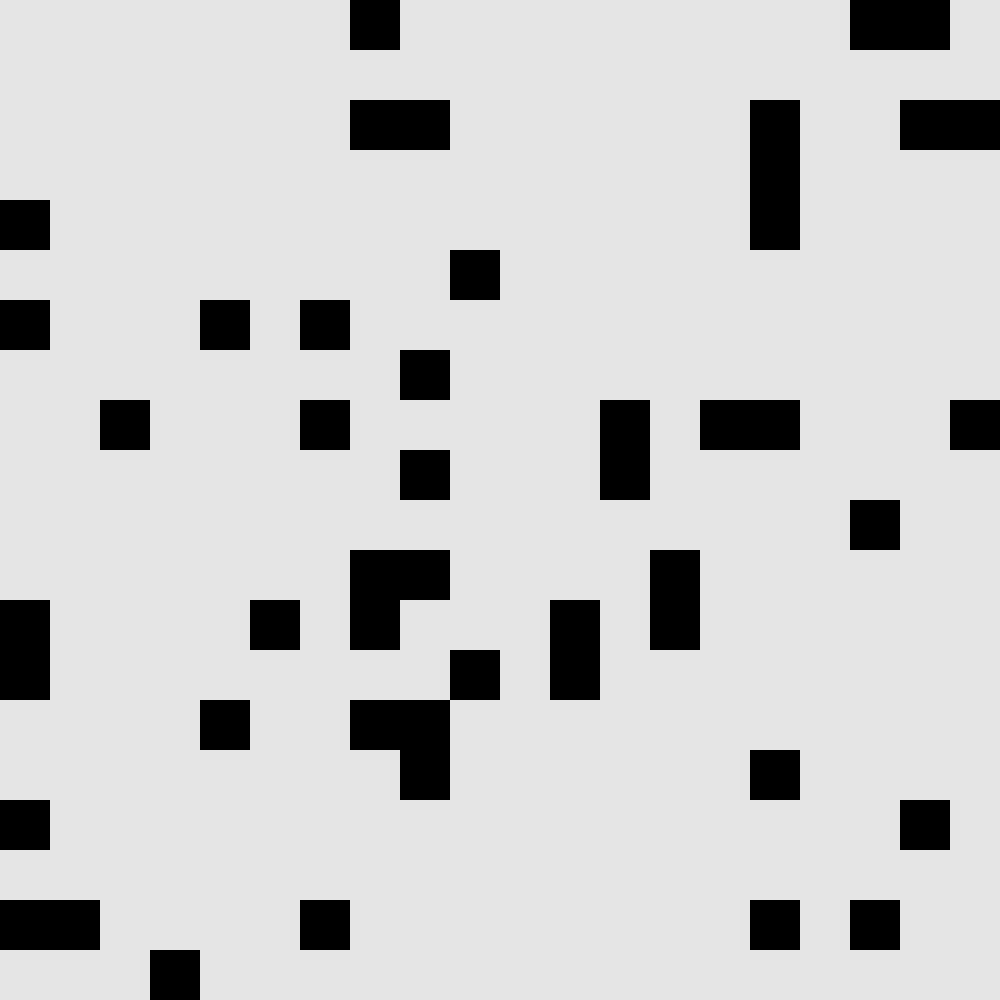}\\
        {\scriptsize random 10\%}

    \end{minipage}
} & \multirow{5}{0.08\linewidth}{\centering 25}    & \multirow{5}{0.08\linewidth}{\centering 27}  &   &   & \underline{Width-0}   & \underline{Width-2}  & \underline{Width-5}\\
& &  & 14  & 12  & 12  & 7  & 1\\
& &  & 6, 7, 10  & 0, 6, 10  & \multirow{3}{0.13\linewidth}{\centering [6; 8; 8; 17; 9; 13; 10; 1; 3; 16; 5; 1]}  & \multirow{3}{0.13\linewidth}{\centering [3; 2; 4; 3; 2; 5; 11]}   & \multirow{3}{0.13\linewidth}{\centering[2]} \\
& &  & 13, 10, 4 & 0, 9, 4  &   &   & \vspace{1.5mm} \\
        
        \hline 

        \multirow{4}{0.12\linewidth}{  
        \begin{minipage}{\linewidth}\vspace{-5mm}
        \centering
        \includegraphics[width=0.6\linewidth]{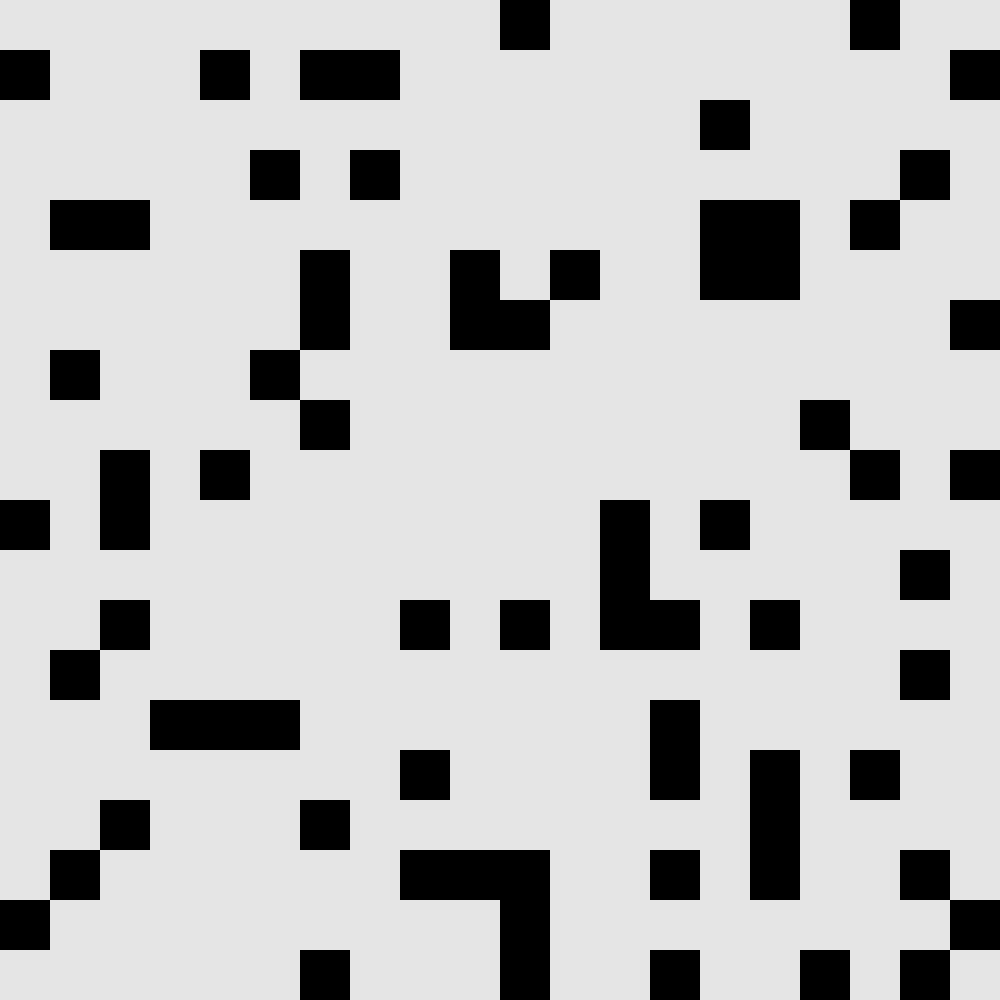}\\
        {\scriptsize random 20\%}
        \end{minipage}}       
        & 
        \multirow{4}{0.08\linewidth}{\centering 25}  & \multirow{4}{0.08\linewidth}{\centering 27}  &  &  & \underline{Width-0}  & \underline{Width-2} & \underline{Width-5}\\
        &  &  & 6  & 3  & 3& 0 & 0\\
        &  &  & 4, 4, 9 & 0, 4, 7&  [1; 1; 1] &     & \\
        &  &  & 2, 3, 7  & 0, 3, 5  &  &    & \vspace{-2mm} \\ 

        \hline
        \multirow{7}{0.12\linewidth}{
            \begin{minipage}{\linewidth} 
                \centering
                \includegraphics[width=0.6\linewidth]{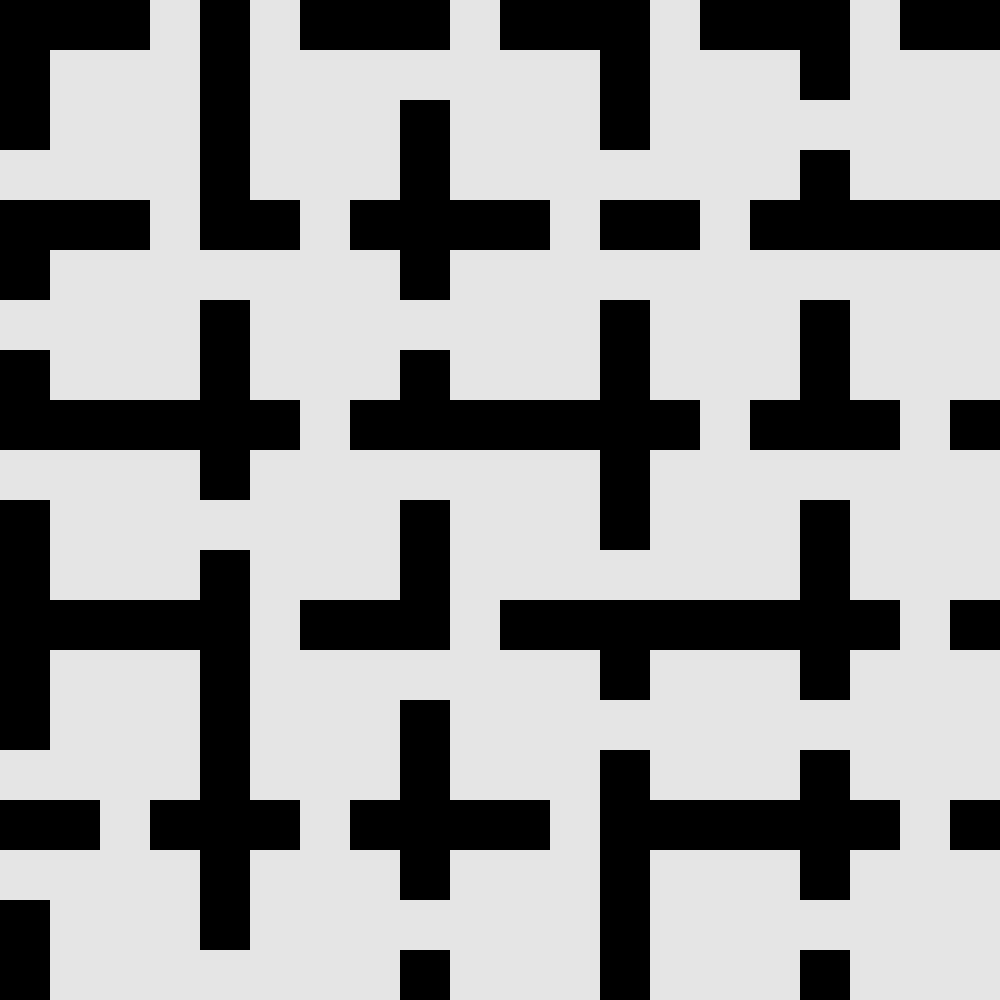}\\
                {\scriptsize room}
        
            \end{minipage} 
        }       & \multirow{7}{0.08\linewidth}{\centering 25}   & \multirow{7}{0.08\linewidth}{\centering 35}   &  &  &  \underline{Width-0}& \underline{Width-2}  & \underline{Width-5}\\
       &  &  &  &  & 19        & 11      & 4\\
       &  &  &20 & 19& \multirow{5}{0.13\linewidth}{\centering [9; 15; 4; 7; 5; 26; 1; 3; 7; 3; 27; 15; 5; 16; 6; 3; 14; 8; 7]}  & \multirow{5}{0.13\linewidth}{\centering [3; 6; 2; 22; 20; 11; 1; 6; 7; 6; 1]}     & \multirow{5}{0.13\linewidth}{\centering [13; 13; 4; 1]}\\
       &  &  & 19, 20, 20 & 0, 19, 19 &  &    & \\
       &  &  & 19, 19, 19 & 0, 19, 19  &  &    & \\
       &  &  &  &    &  &    & \\ 
     
\botline 
\end{tabular}}
\end{table}

The columns \#Plan Changes and \#Path Changes show the number of agents that have a different plan and path than their initial solution. A change in a plan is either a change in the time of visiting a location or changing the visited location completely. Here, we consider a path change as visiting a different vertex that does not exist in the original plan.  For each instance, the first line in these columns show the number of agents for Replan-All, the second lines show the number of agents for TC with widths 0, 2, and~5, respectively and the third lines show the same information for TG.   

For instance, for the first instance (random~10\%), according a \dmapf\ solution computed with Replan-All, 12 agents had to change their initial paths, and 14 agents had to change their initial plans. In a \dmapf\ solution computed with TC: no agent changes its path if the tunnel width is given as 0, while 6 agents change their plans (i.e., traversals of their paths); 6~agents change their paths if the tunnel width is given as 2, while 7 agents change their plans; and 10 agents change their paths if the tunnel width is given as 5, while 10 agents change their plans. 

The last three columns show the number of agents that visit a vertex outside of their tunnels of width 0, 2, and 5, respectively, in a \dmapf solution computed with Replan-All. The square-bracketed lists show how many such vertices outside of tunnels are visited. 

For instance, for the first instance (random~10\%), in a \dmapf\ solution computed with Replan-All, out of the 12 agents who diverged from their paths, 7 agents went outside of their tunnels of width 2, and 1 of them went further away from its path (i.e., outside of its tunnel of width 5).   

\smallskip \noindent{\it Observations and discussions about the quality of solutions.} 
We observe the same amount of increase in the makespan of plans, with all methods. We observe a higher increase in the makespan in the \textit{room} environment, compared to the other environments: all 20 of the existing agents have changed their plans and all except one of the agents have changed their paths. Such higher increases in the makespan (compared to the other two environments) is due to the constrained structure of the environment.  

In terms of divergences; we see that, for every tunnel width, the number of vertices visited outside of the agents tunnels are high for some agents. For instance, for \textit{room} with tunnel width 0, one of the agents visits 27 vertices outside its tunnel (in this case its path). Since the makespan of \dmapf\ plan is 35,  
we can see that the agent diverts from its original path in at least 77\% of its new path. 

These divergence results demonstrate agents significantly changing their plans with Replan-All. Recall that such significant changes in plans are not desired in real-world applications where robots collaborate with humans, and this 
was our motivation for introducing the concept of tunnels. 


 \begin{table}[t] 
    \centering 
    \caption{Average run times of 150 \dmapf instances for different~environments/methods}
    \label{tab:mapf_benchmarks_times_1} 
    \begin{tabular}{cccccccc} 
    \topline 
         Grid & Replan-All & TC\_W0 & TC\_W2 & TC\_W5 & TG\_W0 & TG\_W2 & TG\_W5  \\ 
         \hline
         random 10\% & 39.65 & 47.82 & 44.62 & 43.22 & 48.44 & 48.53 & 48.99 \\
         random 20\% & 33.90  & 39.45 & 37.11 & 36.06 & 41.17 & 41.60  & 42.19 \\
         room        & 38.61 & 39.65 & 34.44 & 34.22 & 42.72 & 41.98 & 43.93 
    \botline
    \end{tabular} 
\end{table} 
  
 \smallskip \noindent{\it Effect of obstacles in computation times.} Table~\ref{tab:mapf_benchmarks_times_1} shows the average of the total computation times for all \dmapf instances in the second set of experiments. When we compare computation times in different environments, we observe that \textit{random~20\%} has the fastest computation times. This is an expected result due to the environment having fewer empty cells for the agents to move; hence, resulting in a smaller problem size. On the other hand, while \textit{room} environment has even fewer empty cells, the computation times are higher due to the tight passages between the rooms. For TC, we observe a similar pattern with the first set of experiments: as the tunnel size increases, the computation becomes faster. For TG, we also observe similar behavior with the first set of experiments: increasing the tunnel size increases the computation times. 

\section{Conclusion}
\dmapf problem considers \mapf problem in dynamic environments where changes can occur in the environment or the team of agents. In this study, we introduced a general definition for \dmapf problem that covers different assumptions on appearance/disappearance of agents and different objective functions studied in the literature, as well as the possible changes that can occur in the environment. We introduced a framework for solving \dmapf that can use different methods for computing the new solutions after changes in the environment, utilizing multi-shot ASP. In addition to formulating the existing approaches Replan-All and Revise-and-Augment using multi-shot ASP, we introduced a new method called Revise-and-Augment-in-Tunnels that combines the advantages of multi-shot solving and re-using the existing solutions. We integrated this method into our framework, and observed its usefulness in terms of computational performance and solution quality. 

\paragraph{\bf Acknowledgments}
We would like to thank Volkan Patoglu for the useful discussions about the real-world applications of \mapf, as well as for the helpful feedback and support. We also thank the anonymous reviewers for their valuable comments.

\bibliographystyle{tlplike}
\bibliography{dmapf_tplp2025_v6}

\appendix

\section{Different cost fuctions} 
\label{app:cost}

The \mapf\ definition introduced in Section~\ref{sec:prel} is more general than the earlier definitions, as it allows us to explicitly state our assumptions. For instance, consider the following two possible behaviours of agents once they reach their goals: 1) every agent waits at its goal until the traversals of all agents are completed (as in our earlier studies), or 2) every agent disappears from the environment (as in \cite{svancara2019}).  Both cases can be expressed within this \mapf\ definition. Suppose that $reach_i$ ($x_i \leqs reach_i \leqs y_i$) denotes the time step at which an agent $a_i$ reaches its goal. If we assume that the agent $a_i$ disappears when it reaches its goal, then $y_i\eqs reach_i$. If we assume that the agent $a_i$ waits at goal until all traversals end, then $reach_{i}$ is the time step where the agent reaches its goal and stays there. 

Similarly, there can be different behaviours of agents at their initial location: 1) every agent appears at their initial locations, at the time step the agent joins the team (as in our earlier studies), or 2) the agents that are joining the team are allowed to wait outside the environment until their initial locations are unoccupied (as in \cite{svancara2019}). In the former case, for agent $a_i$, the starting time $x_i$ of its traversal is the same as its joining time $join_{i}$.
In the latter case, for agent $a_i$, the joining time $join_i$ is the time at which the agent joins the team and the starting time $x_i$ of its traversal is the time at which the agent enters the environment.



The definition of \mapf also allows different cost functions depending on the needs of the particular application.



To define the different cost functions for a \mapf solution $\mathbf{r}_A^\mathbf{P}$, we start by defining the costs of moving from a vertex to a different vertex and waiting at a vertex at a time step of a traversal by an agent $a_i$.

Cost of waiting at a vertex at a time step $t$ for agent $a_i$ with traversal $\trav{i}$ is defined as:
\[cost_w(t) \eqs \begin{cases}

    1 & \text{if } f_i(t) \eqs f_i(t\pluss 1) \text{ for } x_i \leqs t\lts y_i \\
    0 & \text{otherwise}

\end{cases}\]

Cost of moving from a vertex to a different vertex at a time step $t$ for some agent $a_i$ with traversal $\trav{i}$ is defined as:
\[cost_m(t) \eqs \begin{cases}

    1 & \text{if } f_i(t) \neqs f_i(t\pluss 1) \text{ for } x_i \leqs t\lts y_i \\
    0 & \text{otherwise}

\end{cases}\]

The following three functions provide different costs of a single agent's traversal.

\noindent For every agent, the time steps where the agent is waiting at its goal are part of its traversal, since as long as an agent is present in the environment, it can collide with other agents. Therefore, we define the length of a traversal $\trav{i}$ by an agent $a_i$ as:
\begin{equation*}
    cost_{L}(r_{a_i}^{P_i}) \eqs \sum_{t\eqs x_i}^{y_i} cost_w(t) \pluss cost_m(t)
\end{equation*}

Once an agent reaches its goal and stays there until the end of its traversal, waiting at the goal location does not have a cost in terms of completion of the agent's task, since reaching its goal location means that the agent has completed its task. So, we define the cost of a traversal $\trav{i}$ by an agent $a_i$ as:
\begin{equation*}
    cost_{T}(r_{a_i}^{P_i}) \eqs \sum_{t\eqs x_i}^{reach_{i}} cost_w(t) \pluss cost_m(t)
\end{equation*}

Note that if the agents disappear after reaching their goals, $reach_i$ would be the last time step at which the agent $a_i$ is present in the environment, that is $y_i$. Therefore when disappearance is allowed, $cost_{L}(r_{a_i}^{P_i}) \eqs cost_{T}(r_{a_i}^{P_i})$.

When the distance traveled by an agent is important rather than the time spent by the agent, we need to consider the agent's path instead of its traversal. We can also think about this as the total number of time steps where the agent moves to a different vertex. We define the cost of a path $P_i$ with traversal $\trav{i}$ by an agent $a_i$ as:
\begin{equation*}
    cost_{P}(r_{a_i}^{P_i}) \eqs \sum_{t\eqs x_i}^{y_i} cost_m(t)
\end{equation*}

\noindent Note that this is same as the length of the path traversed by the agent $a_i$, which is $|P_i|$. Also, there is no difference between using $y_i$ or $reach_i$ for this function, because after time $reach_i$ the agent does not move to a different vertex until time $y_i$.

Now we define some possible cost functions that can be used for a \mapf solution, using $cost_{L}$, $cost_{T}$ and $cost_{P}$ functions.

\noindent The total time spent by all of the agents until they reach their goals is called the sum of costs of a \mapf plan $\mathbf{r}_A^\mathbf{P}$.

\noindent Sum of costs of a \mapf plan $\mathbf{r}_A^\mathbf{P}$ is defined as:
\begin{equation*}
    cost_{SOC}(\mathbf{r}_A^\mathbf{P}) \eqs \sum_{a_i\ins A} cost_T(a_i)
\end{equation*}

Since the agents are allowed to have cycles in their paths, they may move redundantly in the environment. In this case, we may be more interested in the distance traveled by the agents than the total time. To compute the total distance traveled by the agents, we consider the sum of path lengths of a \mapf plan $\mathbf{r}_A^\mathbf{P}$.

Sum of path lengths of a \mapf plan $\mathbf{r}_A^\mathbf{P}$ is defined as:
\begin{equation*}
    cost_{SOP}(\mathbf{r}_A^\mathbf{P}) \eqs \sum_{a_i\ins A} cost_P(a_i)
\end{equation*}


In another scenario, we might be interested in the agents finishing all tasks as early as possible. In this case, we minimize the makespan of a \mapf plan. The makespan of a \mapf plan $\mathbf{r}_A^\mathbf{P}$ is the time step where all agents in $A$ reach their goals.

Makespan of a \mapf plan $\mathbf{r}_A^\mathbf{P}$ is defined as:
\begin{equation*}
    cost_{M}(\mathbf{r}_A^\mathbf{P})  \eqs max(cost_T(a_i))
\end{equation*}


The $cost$ function of a \mapf instance $\seq{A,G,O,cost,\tau}$ can be one of $cost_{SOC}$, $cost_{SOP}$ and $cost_M$ and they can be extended further if needed by the application.

\section{ASP Formulations}
\label{app:asp_formulations}

Our multi-shot ASP program for \mapf consists of three subprograms \verb|base|, \verb|step| and \verb|check|. In these subprograms,
\begin{itemize}
    \item \verb|vertex(X)| represents a vertex \verb|X| and \verb|edge(X,Y)| represents an edge from vertex \verb|X| to vertex \verb|Y| of the environment,
    \item \verb|agent(A)| represents an agent \verb|A|,
    \item \verb|time(T)| represents a time step \verb|T|,
    \item \verb|init(A,X)| represents the initial location of agent \verb|A| being vertex \verb|X| and similarly \verb|goal(A,Y)| represents the goal location of agent \verb|A| being vertex \verb|Y|,
    \item \verb|plan(A,T,X)| represents the traversal of agent \verb|A| at time step \verb|T|, such that the agent is located at vertex \verb|X| at time step \verb|T|.
\end{itemize}

The \verb|base| program contains an instance specified by \verb|vertex/1|, \verb|edge/2|, \verb|agent/1|, \verb|init/2|, \verb|goal/2| and the following lines specifying the initial time step and the initial step of the traversal, which is the base case of our recursive plan definition:
\begin{verbatim}
#program base.
time(0).

% agent A is at its initial location X at time 0
plan(A,0,X):- init(A,X), agent(A).
\end{verbatim}

The \verb|step| program, with parameter \verb|t| that generates the plan and adds the constraints for time \verb|t| is as follows:
\begin{verbatim}
#program step(t).
time(t).

% agent A can stay at its current location or
% move to an adjacent vertex at time t
{plan(A,t,X)}1 :- plan(A,t-1,X), agent(A).
{plan(A,t,Y): edge(X,Y)}1 :- plan(A,t-1,X), agent(A).

% each agent has exactly one location at time t
:- {plan(A,t,X): vertex(X)}0, agent(A).
:- 2{plan(A,t,X): vertex(X)}, agent(A).

% no two agents are at the same vertex at time t.
:- 2{plan(A,t,X): agent(A)}, vertex(X), time(t).

% no two agents can swap their locations between times t-1 and t
plan_to(X,Y,t,0) :- plan(A,t-1,X), plan(A,t,Y), edge(X,Y), agent(A).
:- plan_to(X,Y,t,_), plan_to(Y,X,t,_).

% if an agent visits its goal at t, then that agent has reached its goal
goal_reached(A,t) :- goal(A,X), plan(A,t,X), agent(A).

% if an agent has reached its goal at t-1, then it reached its goal at t
goal_reached(A,t) :- goal_reached(A,t-1).
\end{verbatim}

The \verb|check| program that checks whether all agents reach their goals until time \verb|t| is as follows:
\begin{verbatim}
#program check(t).
#external query(t).

% there cannot be an agent that did not reach its goal until time t
:- not goal_reached(A,t), agent(A), query(t).

% last step of the plan of each agent is its goal,
% allowing the agents to leave their goals and come back at the end
:- not goal(A,X), plan(A,t,X), agent(A), query(t).
\end{verbatim}

Note that the external atom \verb|query(t)| is only added for time \verb|t| and removed in the next step, allowing the rules in the check program to be added at time \verb|t| and removed at the next time step.

\begin{figure}[ht]
    \centering
    \includegraphics[width=\textwidth]{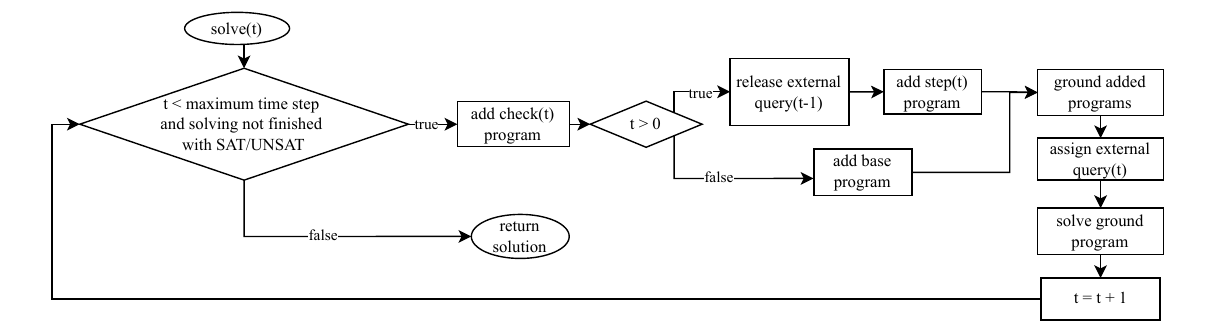}
    \caption{The steps followed for multi-shot solving. This procedure is used for solving \mapf and \dmapf with Replan-All with different values for $t$. We assume the maximum makespan is defined outside of the function.}
    \label{fig:solve_fnc}
\end{figure}

These programs are used in a solving procedure, visualized in Figure~\ref{fig:solve_fnc}. 
Note that this procedure can be used with different values for \verb|t| initially, depending on it being used for ``Solve \mapf'' or ``Solve \dmapf'' blocks of our architecture.

The \verb|newAgent| program that is added for each new agent when their traversals start is as follows:
\begin{verbatim}
#program newAgent(a,i,g,k).
agent(a).
init(a,i). % initial location of a
goal(a,g). % goal location of a

% base case of the recursive plan definition, initial step starts from time k
plan(a,k,X) :- init(a,X), agent(a).

% recursive plan generation until the current makespan for agent a,
% time(T) is defined up to makespan in the existing ground program
{plan(a,T,X)}1 :- plan(a,T-1,X), agent(a), time(T), T >= k.
{plan(a,T,Y): edge(X,Y)}1 :- plan(a,T-1,X), agent(a), time(T), T >= k.

% the agent a is at exactly one vertex at each time step
:- {plan(a,T,X): vertex(X)}0, agent(a), time(T), T >= k.
:- 2{plan(a,T,X): vertex(X)}, agent(a), time(T), T >= k.

% no two agents are at the same vertex at time T. (includes all agents)
:- 2{plan(A,T,X): agent(A)}, vertex(X), time(T), T >= k.

% no swapping between times T-1 and T (includes all agents)
plan_to(X,Y,T,k) :- plan(a,T-1,X), plan(a,T,Y), edge(X,Y), agent(a),
                    time(T), T>=k.
:- plan_to(X,Y,T,_), plan_to(Y,X,T,_), time(T), T>=k.

% agent a has reached its goal until time T if it visits its goal earlier
goal_reached(a,T) :- goal(a,X), plan(a,T,X), agent(a), time(T), T>=k.
goal_reached(a,T) :- goal_reached(a,T-1), time(T), T>=k.

% checks whether a reached its goal until the current time
% similar to check program
:- not goal_reached(a,T), agent(a), query(T).
:- not goal(a,X), plan(a,T,X), agent(a), query(T).
\end{verbatim}

The parameters of the \verb|newAgent| program are the agent \verb|a|, its initial location \verb|i|, its goal location \verb|g| and the starting time of its traversal \verb|k|. Since \verb|time/1| is defined until the current makespan in the existing ground program before adding the \verb|newAgent| program, this program computes possible traversals for agent \verb|a| from time \verb|k| until the current makespan of the existing plan.

For Revise-and-Augment, we have another program to keep the path of the existing agent the same. This program is called \verb|path_constraints| and has two parameters \verb|t|, denoting the current time step of the computation and \verb|k|, denoting the time of last change in the environment. In this program, we extend our signature with the following:
\begin{itemize}
    \item \verb|old_agent(A)| represents that agent \verb|A| exists in the environment before the change is observed. These atoms are added by the controller program.
    
    \item \verb|path(A,X)| represents vertex \verb|X| is in agent \verb|A|'s path. These atoms are added by the controller program, with the information extracted from the existing solution.
    \item \verb|path_pair_count(A,X,Y,C)| represents vertex \verb|X| and vertex \verb|Y|, where \verb|X| and \verb|Y| are different, appear consecutively in agent \verb|A|'s path, \verb|C| times. These atoms are added by the controller program.
    
\end{itemize}

\begin{verbatim}
#program path_constraints(t,k).
#external path_query(t,k). % rescheduling of the plan starts at time k

% every two consecutive and different vertex pair in the plan forms a plan pair
plan_pair(A,X,Y,t,k) :- plan(A,T-1,X), plan(A,T,Y), X!=Y, 
                         old_agent(A), path_query(t,k). 

% number of occurrences of each pair in the plan are counted
plan_pair_count(A,X,Y,C1,t,k):- C1 = #count{X,Y,T: plan(A,T-1,X), plan(A,T,Y),
                                 X!=Y}, plan_pair(A,X,Y,t,k), old_agent(A), 
                                  path_query(t,k).

% the number of occurrences of a pair in the plan and 
% the number of occurrences of a pair in the path should be the same
:- plan_pair_count(A,X,Y,C1,t,k), path_pair_count(A,X,Y,C), C1!=C, 
    old_agent(A), path_query(t,k).

% every vertex appearing in the plan should also appear in the path 
:- plan(A,_,X), not path(A,X,_), old_agent(A), path_query(t,k). 

% every vertex appearing in the path should also appear in the plan 
:- not plan(A,_,X), path(A,X,_), old_agent(A), path_query(t,k). 
\end{verbatim}

For Revise-and-Augment-in-Tunnels, there are two approaches we use when solving this problem with multi-shot ASP. 

The first approach generates the plan atoms using the external \verb|tunnel(A,X)| atoms, denoting that vertex \verb|X| is included in the tunnel of agent \verb|A|. For this approach, the following plan generation rules from the \verb|step| and \verb|newAgent| programs:
\begin{verbatim}
% from step(t) program
{plan(A,t,Y): edge(X,Y)}1 :- plan(A,t-1,X), agent(A).

% from newAgent(a,i,g,k) program
{plan(a,T,Y): edge(X,Y)}1 :- plan(a,T-1,X), agent(a), time(T), T >= k.
\end{verbatim}

\noindent are replaced with:
\begin{verbatim}
% in step(t) program
{plan(A,t,Y): edge(X,Y), tunnel(A,Y)}1 :- plan(A,t-1,X), agent(A).

% in newAgent(a,i,g,k) program
{plan(a,T,Y): edge(X,Y), tunnel(A,Y)}1 :- plan(a,T-1,X), agent(a), time(T), 
                                        T >= k.
\end{verbatim}

The second approach adds a program called \verb|forbidden_locations| after each change in the environment. Here, we use the \verb|forbidden(A,X)| for denoting a location \verb|X| outside of the tunnel of agent \verb|A|. This program is as follows:

\begin{verbatim}
#program forbidden_locations.

% agent A cannot visit location X outside its tunnel
:-  plan(A,T,X), forbidden(A,X), agent(A), time(T).
\end{verbatim}
 
\newpage

\section{Experimental Results for Empty Grids}
\label{app:empty_grid_results}

\begin{table}[h]
    \centering\vspace{-\baselineskip}
    \caption{Experimental results with empty grids (Group 1). Each instance has the same initial and goal locations for the agents. Instances 1 and 2 were run with the \mapf instance Base-1; Instances 3--5 were run with the \mapf instance Base-2. } \vspace{-\baselineskip}
    \label{tab:our_instances1}
    \tablefont\resizebox{.9125\columnwidth}{!}{\begin{tabular}{@{\extracolsep{\fill}}ccccccccc}
    \topline
    & & & Replan& R\&A & TG\_W0 & TG\_W20 & TC\_W0 & TC\_W20 \\
    \hline
    & & Mkspn& \multicolumn{6}{c}{Time (secs): total(top), solve(mid), ground(bot)} \\
    \hline
     \multirow{3}{0.11\linewidth}{\centering Base-1}&  \multirow{3}{0.11\linewidth}{\centering k=0 20~agents}&  \multirow{3}{0.11\linewidth}{\centering 36} & 34.45& 32.92& 46.92& 46.83& 32.74& 32.54 \\
     & & & 4.9& 3.64& 7.99& 7.98& 3.62& 3.58 \\
     & & & 29.55& 29.28& 38.93& 38.85& 29.12& 28.96
     \vspace{-1mm} \\
     \hline
     \multirow{3}{0.11\linewidth}{\centering Instance-1}&  \multirow{3}{0.11\linewidth}{\centering k=1 5~agents}&  \multirow{3}{0.11\linewidth}{\centering 39}& 15.61& 371.36& 18.03& 19.95& 38.05& 21.48 \\
     & & & 0.79& 4.48& 0.41& 1.19& 0.42& 0.87 \\
     & & & 14.82& 366.88& 17.62& 18.76& 37.63& 20.61
     \vspace{-1mm} \\
     \hline
     \multirow{8}{0.11\linewidth}{\centering Instance-2 }&  \multirow{3}{0.11\linewidth}{\centering k=1 2~agents}&  \multirow{3}{0.11\linewidth}{\centering 36}& 4.28& 131.93& 7.2& 5.32& 27.91& 10.23 \\
     & & & 0.49& 3.09& 0.25& 0.74& 0.18& 0.49 \\
     & & & 3.78& 128.84& 6.95& 4.58& 27.73& 9.74
     \vspace{2mm}
     \\
    &  \multirow{3}{0.11\linewidth}{\centering k=3 3~agents}&  \multirow{3}{0.11\linewidth}{\centering 41}& 17.71& 685.92& 15.46& 22.84& 15.59& 19.29 \\
    & & & 1.44& 7.9& 0.5& 1.47& 0.24& 0.66 \\
    & & & 16.26& 678.01& 14.96& 21.37& 15.35& 18.62
    \vspace{-1mm} \\
    \hline
     \multirow{3}{0.11\linewidth}{\centering Base-2}&  \multirow{3}{0.11\linewidth}{\centering k=0 30~agents}&  \multirow{3}{0.11\linewidth}{\centering 36}& 57.35& 55.42& 67.4& 67.3& 55.44& 54.98 \\
     & & & 20.48& 18.65& 17.21& 17.16& 18.52& 18.48 \\
     & & & 36.87& 36.76& 50.19& 50.14& 36.92& 36.5
     \vspace{-1mm} \\
     \hline
     \multirow{3}{0.11\linewidth}{\centering Instance-3}&  \multirow{3}{0.11\linewidth}{\centering k=1 5~agents}&  \multirow{3}{0.11\linewidth}{\centering 39}& 19.01& 764.92& 22.1& 28.67& 61.89& 29.77 \\
     & & & 1.15& 31.56& 0.38& 5.72& 0.35& 1.02 \\
     & & & 17.87& 733.37& 21.72& 22.95& 61.54& 28.75
     \vspace{-1mm} \\
     \hline
     \multirow{3}{0.11\linewidth}{\centering Instance-4}&  \multirow{3}{0.11\linewidth}{\centering k=1 10~agents}&  \multirow{3}{0.11\linewidth}{\centering 39}& 28.96& 757.6& 35.09& 36.9& 72.59& 39.81 \\
     & & & 1.07& 15.89& 1.24& 2.5& 0.72& 1.44 \\
     & & & 27.88& 741.71& 33.84& 34.4& 29.12& 38.37
     \vspace{-1mm} \\
     \hline
     \multirow{17}{0.11\linewidth}{\centering Instance-5}&  \multirow{3}{0.11\linewidth}{\centering k=1 2~agents}&  \multirow{3}{0.11\linewidth}{\centering 39}& 12.89& - & 14.7& 17.68& 55.53& 24.21 \\
     & & & 0.73& - & 0.24& 1.48& 0.31& 0.71 \\
     & & & 12.16& - & 14.46& 16.19& 55.22& 23.5
     \vspace{2mm}
     \\
    &  \multirow{3}{0.11\linewidth}{\centering k=2 2~agents}&  \multirow{3}{0.11\linewidth}{\centering 40}& 9.12& - & 8.24& 11.59& 9.64& 12.18 \\
    & & & 0.7& - & 0.22& 1.13& 0.21& 0.81 \\
    & & & 8.42& - & 8.01& 10.46& 9.43& 11.37
    \vspace{2mm}
    \\
    &  \multirow{3}{0.11\linewidth}{\centering k=3 2~agents}& \multirow{3}{0.11\linewidth}{\centering 40} &6.44& - & 6.5& 8.31& 8.06& 11 \\
    & & & 0.75& - & 0.35& 1.19& 0.26& 2.25 \\
    & & & 5.69& - & 6.14& 7.12& 7.8& 8.75
    \vspace{2mm}
    \\
    &  \multirow{3}{0.11\linewidth}{\centering k=4 2~agents}&  \multirow{3}{0.11\linewidth}{\centering 40}&6.97& - & 6.47& 9.03& 8.44& 9.98 \\
    & & & 1& - & 0.29& 1.33& 0.24& 0.93 \\
    & & & 5.97& - & 6.18& 7.7& 8.2& 9.05
    \vspace{2mm}
    \\
    &  \multirow{3}{0.11\linewidth}{\centering k=5 2~agents}&  \multirow{3}{0.11\linewidth}{\centering 41}& 10.61& - & 8.78& 14.21& 10.68& 13.94 \\
    & & & 0.88& - & 0.28& 1.41& 0.25& 0.99 \\
    & & & 9.73& - & 8.5& 12.8& 10.43& 12.95
    \botline
    \vspace{-8mm}
    \end{tabular}}
\end{table}

\begin{table}[ht]
    \centering
    \caption{Experimental results with empty grids (Group 2). Each instance has the same initial and goal locations for the agents. Instances 1 and 2 were run with the \mapf instance Base-1; Instances 3--5 were run with the \mapf instance Base-2. } \vspace{-\baselineskip}
    \label{tab:our_instances2}
    \tablefont\resizebox{.9125\columnwidth}{!}{\begin{tabular}{@{\extracolsep{\fill}}cccccccc}
    \topline
& & & Replan& TG\_W0& TG\_W20& TC\_W0& TC\_W20 \vspace{-1mm} \\
     \hline 
& & Mkspn& \multicolumn{5}{c}{Time (secs): total(top), solve(mid), ground(bot)} \\
\hline
 \multirow{3}{0.11\linewidth}{\centering Base-1}&  \multirow{3}{0.11\linewidth}{\centering k=0 20~agents}&  \multirow{3}{0.11\linewidth}{\centering 34} & 27.68& 38.61& 38.49& 27.61& 27.48 \\
&& & 2.07& 5.32& 5.3& 2.31& 2.32 \\
&& & 25.6& 33.29& 33.19& 25.3& 25.16     \vspace{-1mm} \\
     \hline
 \multirow{3}{0.11\linewidth}{\centering Instance-1}&  \multirow{3}{0.11\linewidth}{\centering k=1 5~agents}&  \multirow{3}{0.11\linewidth}{\centering 39} & 16.67& 18.25& 22.57& 32.44& 21.26 \\
&& & 0.98& 0.71& 1.84& 0.22& 1.02 \\
&& & 15.7& 17.54& 20.73& 32.22& 20.24     \vspace{-1mm} \\
     \hline
 \multirow{8}{0.11\linewidth}{\centering Instance-2 }&  \multirow{3}{0.11\linewidth}{\centering k=1 2~agents}&  \multirow{3}{0.11\linewidth}{\centering 37} & 7.89& 9.67& 11.11& 24.89& 13.02 \\
&& & 0.44& 0.14& 0.73& 0.12& 0.48 \\
&& & 7.45& 9.52& 10.38& 24.77& 12.53       \vspace{2mm}\\
&  \multirow{3}{0.11\linewidth}{\centering k=3 3~agents}&  \multirow{3}{0.11\linewidth}{\centering 41} & 13.58& 11.83& 18.19& 9.76& 15.32 \\
&& & 0.65& 0.41& 1.03& 0.18& 0.58 \\
&& & 12.93& 11.42& 17.16& 9.58& 14.75     \vspace{-1mm} \\
     \hline
 \multirow{3}{0.11\linewidth}{\centering Base-2}&  \multirow{3}{0.11\linewidth}{\centering k=0 30~agents}&  \multirow{3}{0.11\linewidth}{\centering 34} & 48.88& 61.69& 61.43& 48.59& 48.82 \\
&& & 13.67& 16.04& 15.97& 14.69& 14.74 \\
&& & 35.21& 45.65& 45.46& 33.9& 34.08     \vspace{-1mm} \\
     \hline
 \multirow{3}{0.11\linewidth}{\centering Instance-3}&  \multirow{3}{0.11\linewidth}{\centering k=1 5~agents}&  \multirow{3}{0.11\linewidth}{\centering 39} & 25.72& 23.82& 29.58& 57.18& 31.62 \\
&& & 4.86& 0.53& 1.42& 0.56& 1.06 \\
&& & 20.86& 23.28& 28.15& 56.62& 30.56 
\vspace{-1mm} \\
     \hline
 \multirow{3}{0.11\linewidth}{\centering Instance-4}&  \multirow{3}{0.11\linewidth}{\centering k=1 10~agents}&  \multirow{3}{0.11\linewidth}{\centering 39} & 33.38& 36.13& 40.44& 66.37& 41.04 \\
&& & 3.71& 1.51& 1.77& 0.84& 2.62 \\
&& & 29.67& 34.63& 38.67& 25.3& 38.43     \vspace{-1mm} \\
     \hline
 \multirow{17}{0.11\linewidth}{\centering Instance-5}&  \multirow{3}{0.11\linewidth}{\centering k=1 2~agents}&  \multirow{3}{0.11\linewidth}{\centering 37} & 11.48& 13.04& 15.88& 47.75& 20.88 \\
&& & 0.98& 0.24& 1& 0.16& 0.9 \\
&& & 10.5& 12.8& 14.88& 47.59& 19.98          \vspace{2mm} \\

&  \multirow{3}{0.11\linewidth}{\centering k=2 2~agents}&  \multirow{3}{0.11\linewidth}{\centering 38} & 8.08& 5.99& 10.6& 6.25& 12.31 \\
&& & 0.96& 0.19& 1.04& 0.17& 2.38 \\
&& & 7.12& 5.8& 9.56& 6.08& 9.92          \vspace{2mm} \\
 
&  \multirow{3}{0.11\linewidth}{\centering k=3 2~agents}&  \multirow{3}{0.11\linewidth}{\centering 41} & 15.04& 9.71& 19.87& 10.14& 18.14 \\
&& & 0.99& 0.25& 1.36& 0.2& 0.87 \\
&& & 14.06& 9.47& 18.51& 9.93& 17.27          \vspace{2mm} \\

&  \multirow{3}{0.11\linewidth}{\centering k=4 2~agents}&  \multirow{3}{0.11\linewidth}{\centering 41} & 7.16& 4.32& 11.38& 5.65& 9.94 \\
&& & 1.44& 0.24& 3.91& 0.23& 1.48 \\
&& & 5.72& 4.08& 7.48& 5.42& 8.46          \vspace{2mm} \\

&  \multirow{3}{0.11\linewidth}{\centering k=5 2~agents}&  \multirow{3}{0.11\linewidth}{\centering 43} & 15.44& 8.53& 18.66& 9.94& 17.46 \\
&& & 2.28& 0.27& 1.52& 0.42& 0.96 \\
&& & 13.16& 8.25& 17.14& 9.53& 16.5 \\
    \botline
    \end{tabular}}
\end{table}

\begin{table}[ht]
    \centering
    \caption{Experimental results with empty grids (Group 3). Each instance has the same initial and goal locations for the agents. Instances 1 and 2 were run with the \mapf instance Base-1; Instances 3--5 were run with the \mapf instance Base-2. } \vspace{-\baselineskip}
    \label{tab:our_instances3}
    \tablefont\resizebox{.9125\columnwidth}{!}{\begin{tabular}{@{\extracolsep{\fill}}cccccccc}
    \topline
& & & Replan& TG\_W0& TG\_W20& TC\_W0& TC\_W20   \vspace{-1mm} \\
     \hline 
& & Mkspn&       \multicolumn{5}{c}{Time (secs): total(top), solve(mid), ground(bot)}    \vspace{-1mm} \\
     \hline
 \multirow{3}{0.11\linewidth}{\centering Base-1}&  \multirow{3}{0.11\linewidth}{\centering k=0 20~agents}&  \multirow{3}{0.11\linewidth}{\centering 38} & 36.1& 54.5& 54.36& 35.13& 35.58 \\
&& & 3.82& 12.08& 12.09& 3.48& 3.52 \\
&& & 32.28& 42.42& 42.28& 31.65& 32.06         \vspace{-1mm} \\
     \hline
 \multirow{3}{0.11\linewidth}{\centering Instance-1}&  \multirow{3}{0.11\linewidth}{\centering k=1 5~agents}&  \multirow{3}{0.11\linewidth}{\centering 38} & 8.17& 12.56& 11.83& 34.35& 13.53 \\
&& & 0.6& 0.35& 1.93& 0.24& 0.64 \\
&& & 7.57& 12.21& 9.9& 34.11& 12.89         \vspace{-1mm} \\
     \hline
 \multirow{8}{0.11\linewidth}{\centering Instance-2 }&  \multirow{3}{0.11\linewidth}{\centering k=1 2~agents}&  \multirow{3}{0.11\linewidth}{\centering 38} & 3.92& 7.33& 6.88& 30.4& 10.54 \\
&& & 0.48& 0.25& 2.06& 0.17& 0.62 \\
&& & 3.43& 7.08& 4.82& 30.23& 9.93      \vspace{2mm}\\
&  \multirow{3}{0.11\linewidth}{\centering k=3 3~agents}&  \multirow{3}{0.11\linewidth}{\centering 39} & 7.61& 6.59& 10.45& 6.67& 9.29 \\
&& & 0.51& 0.18& 0.85& 0.16& 0.54 \\
&& & 7.11& 6.41& 9.6& 6.51& 8.75         \vspace{-1mm} \\
     \hline
 \multirow{3}{0.11\linewidth}{\centering Base-2}&  \multirow{3}{0.11\linewidth}{\centering k=0 30~agents}&  \multirow{3}{0.11\linewidth}{\centering 38} & 56.53& 81.91& 85.66& 50.46& 50.39 \\
&& & 13.46& 23.34& 26.94& 7.44& 7.49 \\
&& & 43.07& 58.57& 58.72& 43.01& 42.9        \vspace{-1mm} \\
     \hline
 \multirow{3}{0.11\linewidth}{\centering Instance-3}&  \multirow{3}{0.11\linewidth}{\centering k=1 5~agents}&  \multirow{3}{0.11\linewidth}{\centering 38} & 10.58& 16.51& 13.38& 65.11& 19.55 \\
&& & 0.92& 0.39& 1.12& 0.3& 0.87 \\
&& & 9.66& 16.12& 12.26& 64.81& 18.68      \vspace{-1mm} \\
     \hline
 \multirow{3}{0.11\linewidth}{\centering Instance-4}&  \multirow{3}{0.11\linewidth}{\centering k=1 10~agents}&  \multirow{3}{0.11\linewidth}{\centering 38} & 19.59& 27.54& 24.09& 76.88& 27.9 \\
&& & 1.11& 0.66& 1.4& 0.66& 1.03 \\
&& & 18.48& 26.87& 22.7& 31.65& 26.87        \vspace{-1mm} \\
     \hline
 \multirow{17}{0.11\linewidth}{\centering Instance-5}&  \multirow{3}{0.11\linewidth}{\centering k=1 2~agents}&  \multirow{3}{0.11\linewidth}{\centering 38} & 5.37& 10.37& 6.93& 60.24& 14.57 \\
&& & 0.82& 0.3& 0.96& 0.21& 0.79 \\
&& & 4.55& 10.07& 5.97& 60.03& 13.78      \vspace{2mm}\\
&  \multirow{3}{0.11\linewidth}{\centering k=2 2~agents}&  \multirow{3}{0.11\linewidth}{\centering 38} & 7.4& 4.7& 7.07& 5.78& 7.91 \\
&& & 1.12& 0.2& 0.96& 0.19& 0.67 \\
&& & 6.28& 4.5& 6.12& 5.59& 7.23      \vspace{2mm}\\
&  \multirow{3}{0.11\linewidth}{\centering k=3 2~agents}&  \multirow{3}{0.11\linewidth}{\centering 38} & 6.48& 4.55& 7.28& 5.83& 8.32 \\
&& & 0.79& 0.21& 1.05& 0.21& 0.71 \\
&& & 5.7& 4.34& 6.23& 5.62& 7.62     \vspace{2mm} \\
&  \multirow{3}{0.11\linewidth}{\centering k=4 2~agents}&  \multirow{3}{0.11\linewidth}{\centering 38} & 6.05& 4.56& 8.28& 5.66& 8.55 \\
&& & 0.81& 0.24& 1.82& 0.2& 0.73 \\
&& & 5.24& 4.31& 6.46& 5.46& 7.81     \vspace{2mm} \\
&  \multirow{3}{0.11\linewidth}{\centering k=5 2~agents}&  \multirow{3}{0.11\linewidth}{\centering 38} & 6.08& 4.59& 8.27& 5.94& 9.98 \\
&& & 0.81& 0.24& 1.85& 0.22& 1.98 \\
&& & 5.27& 4.34& 6.42& 5.73& 7.99 \\
    \botline
    \end{tabular}}
\end{table}

\begin{table}[ht]
    \centering
    \caption{Experimental results with empty grids (Group 4). Each instance has the same initial and goal locations for the agents. Instances 1 and 2 were run with the \mapf instance Base-1; Instances 3--5 were run with the \mapf instance Base-2. } \vspace{-\baselineskip}
    \label{tab:our_instances4}
    \tablefont\resizebox{.9125\columnwidth}{!}{\begin{tabular}{@{\extracolsep{\fill}}cccccccc}
    \topline
& & & Replan& TG\_W0& TG\_W20& TC\_W0& TC\_W20 \vspace{-1mm} \\
     \hline
& & Mkspn&  \multicolumn{5}{c}{Time (secs): total(top), solve(mid), ground(bot)} \vspace{-1mm} \\
     \hline
 \multirow{3}{0.11\linewidth}{\centering Base-1}&  \multirow{3}{0.11\linewidth}{\centering k=0 20~agents}&  \multirow{3}{0.11\linewidth}{\centering 36} & 41.41& 45.72& 45.78& 34.16& 34.15 \\
&& & 11.52& 6.11& 6.14& 4.53& 4.53 \\
&& & 29.89& 39.6& 39.64& 29.63& 29.62    \vspace{-1mm} \\
     \hline
 \multirow{3}{0.11\linewidth}{\centering Instance-1}&  \multirow{3}{0.11\linewidth}{\centering k=1 5~agents}&  \multirow{3}{0.11\linewidth}{\centering 36} & 7.64& 11.91& 10.46& 31.37& 13.04 \\
&& & 0.5& 0.51& 1.49& 0.25& 0.97 \\
&& & 7.14& 11.4& 8.97& 31.11& 12.08    \vspace{-1mm} \\
     \hline
     \multirow{8}{0.11\linewidth}{\centering Instance-2 }&  \multirow{3}{0.11\linewidth}{\centering k=1 2~agents}&  \multirow{3}{0.11\linewidth}{\centering 36} & 3.68& 6.83& 5.98& 27.64& 9.19 \\
&& & 0.42& 0.2& 1.78& 0.15& 0.51 \\
&& & 3.26& 6.62& 4.2& 27.48& 8.67      \vspace{2mm}\\
&  \multirow{3}{0.11\linewidth}{\centering k=3 3~agents}&  \multirow{3}{0.11\linewidth}{\centering 36} & 5.26& 4.55& 6.65& 4.92& 6.85 \\
&& & 0.51& 0.2& 0.86& 0.2& 0.54 \\
&& & 4.75& 4.35& 5.79& 4.72& 6.31    \vspace{-1mm} \\
     \hline
 \multirow{3}{0.11\linewidth}{\centering Base-2}&  \multirow{3}{0.11\linewidth}{\centering k=0 30~agents}&  \multirow{3}{0.11\linewidth}{\centering 36} & 48.53& 70.63& 71.79& 47.27& 46.8 \\
&& & 8.67& 16.53& 16.6& 7.45& 7.39 \\
&& & 39.86& 54.11& 55.19& 39.83& 39.41    \vspace{-1mm} \\
     \hline
 \multirow{3}{0.11\linewidth}{\centering Instance-3}&  \multirow{3}{0.11\linewidth}{\centering k=1 5~agents}&  \multirow{3}{0.11\linewidth}{\centering 36} & 11.61& 15.66& 13.68& 59.06& 18.73 \\
&& & 1.33& 0.52& 1.43& 0.41& 0.98 \\
&& & 10.29& 15.14& 12.25& 58.66& 17.76   \vspace{-1mm} \\
     \hline
 \multirow{3}{0.11\linewidth}{\centering Instance-4}&  \multirow{3}{0.11\linewidth}{\centering k=1 10~agents}&  \multirow{3}{0.11\linewidth}{\centering 36} & 18.96& 25.58& 26.02& 66.92& 27.42 \\
&& & 1.67& 1.08& 4.31& 0.59& 1.76 \\
&& & 17.28& 24.51& 21.71& 29.63& 25.67    \vspace{-1mm} \\
     \hline
 \multirow{17}{0.11\linewidth}{\centering Instance-5}&  \multirow{3}{0.11\linewidth}{\centering k=1 2~agents}&  \multirow{3}{0.11\linewidth}{\centering 36} & 4.91& 9.82& 7.74& 54.32& 13.61 \\
&& & 0.64& 0.25& 0.9& 0.2& 0.66 \\
&& & 4.26& 9.57& 6.84& 54.12& 12.95      \vspace{2mm}\\
&  \multirow{3}{0.11\linewidth}{\centering k=2 2~agents}&  \multirow{3}{0.11\linewidth}{\centering 36} & 5.04& 4.81& 7.84& 5.34& 7.75 \\
&& & 0.56& 0.19& 0.98& 0.2& 0.63 \\
&& & 4.48& 4.62& 6.86& 5.14& 7.12      \vspace{2mm}\\
&  \multirow{3}{0.11\linewidth}{\centering k=3 2~agents}&  \multirow{3}{0.11\linewidth}{\centering 36} & 5.3& 4.4& 7.99& 5.17& 7.74 \\
&& & 0.62& 0.19& 0.98& 0.2& 0.65 \\
&& & 4.67& 4.21& 7.01& 4.97& 7.09      \vspace{2mm}\\
&  \multirow{3}{0.11\linewidth}{\centering k=4 2~agents}&  \multirow{3}{0.11\linewidth}{\centering 36} & 5.25& 4.04& 8.09& 5.06& 7.89 \\
&& & 0.61& 0.21& 1.04& 0.19& 0.69 \\
&& & 4.64& 3.83& 7.05& 4.88& 7.2      \vspace{2mm}\\
&  \multirow{3}{0.11\linewidth}{\centering k=5 2~agents}&  \multirow{3}{0.11\linewidth}{\centering 39} & 15.51& 10& 21.28& 10.71& 18.06 \\
&& & 1.28& 0.24& 1.34& 0.24& 0.82 \\
&& & 14.23& 9.76& 19.94& 10.46& 17.25 \\
    \botline
    \end{tabular}}
\end{table}

\begin{table}[ht]
    \centering
    \caption{Experimental results with empty grids (Group 5). Each instance has the same initial and goal locations for the agents. Instances 1 and 2 were run with the \mapf instance Base-1; Instances 3--5 were run with the \mapf instance Base-2. } \vspace{-\baselineskip}
    \label{tab:our_instances5}
    \tablefont\resizebox{.9125\columnwidth}{!}{\begin{tabular}{@{\extracolsep{\fill}}cccccccc}
    \topline
& & & Replan& TG\_W0& TG\_W20& TC\_W0& TC\_W20 \vspace{-1mm} \\
     \hline 
& & Mkspn&  \multicolumn{5}{c}{Time (secs): total(top), solve(mid), ground(bot)} \vspace{-1mm} \\
     \hline
 \multirow{3}{0.11\linewidth}{\centering Base-1}&  \multirow{3}{0.11\linewidth}{\centering k=0 20~agents}&  \multirow{3}{0.11\linewidth}{\centering 36} & 31.61& 43.46& 43.28& 32.91& 32.7 \\
&& & 2.38& 4.7& 4.69& 3.87& 3.87 \\
&& & 29.24& 38.76& 38.59& 29.05& 28.83    \vspace{-1mm} \\
     \hline
 \multirow{3}{0.11\linewidth}{\centering Instance-1}&  \multirow{3}{0.11\linewidth}{\centering k=1 5~agents}&  \multirow{3}{0.11\linewidth}{\centering 36} & 7.89& 11.81& 10.14& 31.12& 13.28 \\
&& & 0.54& 0.3& 1& 0.34& 0.58 \\
&& & 7.35& 11.51& 9.14& 30.78& 12.71    \vspace{-1mm} \\
     \hline
 \multirow{8}{0.11\linewidth}{\centering Instance-2 }&  \multirow{3}{0.11\linewidth}{\centering k=1 2~agents}&  \multirow{3}{0.11\linewidth}{\centering 36} & 3.73& 8.09& 5.05& 27.15& 9.69 \\
&& & 0.45& 0.27& 0.76& 0.15& 0.51 \\
&& & 3.29& 7.82& 4.28& 27& 9.18      \vspace{2mm}\\
&  \multirow{3}{0.11\linewidth}{\centering k=3 3~agents}&  \multirow{3}{0.11\linewidth}{\centering 36} & 5.38& 5.78& 6.81& 5.17& 7.03 \\
&& & 0.48& 0.37& 0.88& 0.19& 0.51 \\
&& & 4.89& 5.41& 5.93& 4.98& 6.53    \vspace{-1mm} \\
     \hline
 \multirow{3}{0.11\linewidth}{\centering Base-2}&  \multirow{3}{0.11\linewidth}{\centering k=0 30~agents}&  \multirow{3}{0.11\linewidth}{\centering 38} & 48.87& 77.08& 77.54& 60.77& 60.11 \\
&& & 5.99& 19.27& 19.33& 17.51& 17.37 \\
&& & 42.87& 57.81& 58.21& 43.25& 42.74    \vspace{-1mm} \\
     \hline
 \multirow{3}{0.11\linewidth}{\centering Instance-3}&  \multirow{3}{0.11\linewidth}{\centering k=1 5~agents}&  \multirow{3}{0.11\linewidth}{\centering 38} & 11.73& 16.67& 13.69& 64.12& 20.17 \\
&& & 1& 0.55& 1.3& 0.43& 0.9 \\
&& & 10.73& 16.12& 12.39& 63.69& 19.27   \vspace{-1mm} \\
     \hline
 \multirow{3}{0.11\linewidth}{\centering Instance-4}&  \multirow{3}{0.11\linewidth}{\centering k=1 10~agents}&  \multirow{3}{0.11\linewidth}{\centering 39} & 23.26& 31.13& 30.04& 75.64& 33.85 \\
&& & 0.97& 1.34& 2.82& 0.45& 1.8 \\
&& & 22.29& 29.78& 27.22& 29.05& 32.05   \vspace{-1mm} \\
     \hline
 \multirow{17}{0.11\linewidth}{\centering Instance-5}&  \multirow{3}{0.11\linewidth}{\centering k=1 2~agents}&  \multirow{3}{0.11\linewidth}{\centering 38} & 6.05& 10.11& 7.06& 58.92& 15.05 \\
&& & 0.83& 0.28& 1.06& 0.22& 0.75 \\
&& & 5.22& 9.84& 6.01& 58.7& 14.29      \vspace{2mm}\\
&  \multirow{3}{0.11\linewidth}{\centering k=2 2~agents}&  \multirow{3}{0.11\linewidth}{\centering 38} & 7.54& 4.73& 7.22& 6.09& 8.19 \\
&& & 2.6& 0.19& 0.98& 0.2& 0.7 \\
&& & 4.93& 4.54& 6.24& 5.89& 7.48     \vspace{2mm} \\
&  \multirow{3}{0.11\linewidth}{\centering k=3 2~agents}&  \multirow{3}{0.11\linewidth}{\centering 38} & 5.74& 4.45& 7.28& 5.89& 8.24 \\
&& & 0.73& 0.19& 1.05& 0.21& 0.73 \\
&& & 5.01& 4.26& 6.23& 5.67& 7.51      \vspace{2mm}\\
&  \multirow{3}{0.11\linewidth}{\centering k=4 2~agents}&  \multirow{3}{0.11\linewidth}{\centering 38} & 5.85& 6.45& 7.4& 5.72& 8.53 \\
&& & 0.72& 0.26& 1.09& 0.21& 0.75 \\
&& & 5.12& 6.2& 6.32& 5.51& 7.78      \vspace{2mm}\\
&  \multirow{3}{0.11\linewidth}{\centering k=5 2~agents}&  \multirow{3}{0.11\linewidth}{\centering 43} & 22.96& 13& 30.79& 15.66& 26.11 \\
&& & 0.95& 0.24& 1.57& 0.23& 0.95 \\
&& & 22.01& 12.75& 29.22& 15.44& 25.16 \\
    \botline
    \end{tabular}}
\end{table}

\end{document}